\pdfoutput=1

\documentclass[11pt]{article}

\usepackage[final]{acl}

\usepackage{times}
\usepackage{latexsym}
\usepackage{placeins}
\usepackage[T1]{fontenc}

\usepackage[utf8]{inputenc}

\usepackage{microtype}

\usepackage{inconsolata}

\usepackage{graphicx}
\usepackage{multirow}
\usepackage{tabularx}
\usepackage{booktabs}
\usepackage{enumitem}
\usepackage{amsmath}
\usepackage{siunitx}
\usepackage{pifont}
\newcommand{\cmark}{\ding{51}}%
\newcommand{\xmark}{\ding{55}}%
\usepackage{subcaption}
\usepackage{xspace}
\usepackage{xcolor}

\title{Leveraging Multilingual Training for Authorship Representation: Enhancing Generalization across Languages and Domains}

\newcommand{\myparagraph}[1]{\paragraph{#1}}

\newcommand{\camerareadytext}[1]{#1}

\author{Junghwan Kim, Haotian Zhang \& David Jurgens \\
University of Michigan\\
\texttt{\{kimjhj,zht,jurgens\}@umich.edu}
}

\begin{document}
\maketitle
\begin{abstract}
Authorship representation (AR) learning, which models an author's unique writing style, has demonstrated strong performance in authorship attribution tasks.
However, prior research has primarily focused on monolingual settings---mostly in English---leaving the potential benefits of multilingual AR models underexplored.
We introduce a novel method for multilingual AR learning that incorporates two key innovations: probabilistic content masking, which encourages the model to focus on stylistically indicative words rather than content-specific words, and language-aware batching, which improves contrastive learning by reducing cross-lingual interference.
Our model is trained on over 4.5 million authors across 36 languages and 13 domains.
It consistently outperforms monolingual baselines in 21 out of 22 non-English languages, achieving an average Recall@8 improvement of 4.85\%, with a maximum gain of 15.91\% in a single language.
Furthermore, it exhibits stronger cross-lingual and cross-domain generalization compared to a monolingual model trained solely on English.
Our analysis confirms the effectiveness of both proposed techniques, highlighting their critical roles in the model's improved performance.
\end{abstract}

\section{Introduction}
\label{sec:introduction}
Authorship representation (AR) models~\cite{zhu2021idiosyncratic,riverasoto2021luar,wang2023luar} capture an author's distinctive writing style by encoding documents written by the same author as nearby vectors in the embedding space.
Initially developed for authorship attribution, AR models have since proven useful for a wide range of applications, including machine-generated text detection~\cite{riverasoto2024fewshot}, text style transfer~\cite{horvitz2024tinystyler,liu2024authorship}, stylistic similarity measurement~\cite{alshomary2024latent}, authorship obfuscation~\cite{bao2024keep,fisher2024styleremix}, and personalized text generation~\cite{neelakanteswara2024rags}.


Despite this growing versatility, most prior work on AR has focused exclusively on English, limiting the generalizability of AR models across languages.
As NLP systems are increasingly deployed globally, multilingual support has become critical.
Yet developing effective AR models for other languages remains difficult due to two central challenges: \textit{data scarcity} and \textit{topic dependence}.

The lack of large, diverse author-labeled datasets poses a major challenge for AR modeling in non-English languages.
While English corpora used in \citet{riverasoto2021luar} include up to 1.1 million authors spanning three domains, most existing non-English datasets contain only a few hundred authors from a single domain~\cite{stamatatos2015overview,avram2023bertbased,gabrovsek2023authorship,nitu2024authorship,delanghe2024unsupervised,misini2024automatic,hossain2025authornet}, limiting the feasibility of accurate AR modeling beyond English.

Moreover, AR models often conflate stylistic signals with topic-related features~\cite{sawatphol2022topicreg,wegmann2022cav}, which weakens their ability to generalize across domains.
While some recent methods have attempted to reduce topic bias, they often depend on language-specific tools such as semantic representations~\cite{hu2024contrastdistaa} or syntactic parsers~\cite{wang2023luar}.
However, these tools are rarely available for non-English languages, severely hindering the adaptation of existing AR approaches in multilingual settings.

To address these challenges, we pose the following research questions:
Can multilingual training with a single shared model improve authorship representations in low-resource languages?
What modeling strategies are effective for reducing topic bias and isolating language-agnostic stylistic features in multilingual AR models?

In this paper, we propose a novel multilingual authorship representation method that enables joint training of a single embedding model across multiple languages.
We introduce two innovative techniques to improve the model's robustness to topic shifts and enhance stability during multilingual contrastive learning, without requiring language-specific resources.
The first technique, probabilistic content masking, encourages the model to focus on stylistic cues rather than topic-based ones.
To achieve this, we identify frequently occurring tokens as function words---words more likely to signal stylistic choices---and mask the remaining content tokens randomly.
The second technique, language-aware batching, groups same-language examples into contrastive batches, thereby providing a more informative contrastive objective and greater training efficiency.

Our multilingual authorship representation models are trained on texts from 36 languages spanning 19 language families and 17 distinct script systems, covering over 4.5 million authors.
Our experiments show that these multilingual models consistently outperform monolingual models, particularly in languages with less author-labeled data.
In 21 out of 22 non-English languages, our multilingual model achieves higher Recall@8 than its monolingual counterpart, with an average improvement of 4.85\%.
Languages with limited author-labeled data benefit the most: Recall@8 in Kazakh and Georgian improves by over 15\%.
Moreover, we demonstrate that multilingual training improves authorship attribution performance even in languages and domains not seen during training.

This paper makes four key contributions:
(1) We propose a novel multilingual AR learning method that enables training a single model across multiple languages without relying on language-specific resources.
(2) We demonstrate that multilingual training consistently outperforms monolingual baselines, even in unseen languages and domains.
(3) We conduct a detailed ablation study that highlights the effectiveness of our proposed techniques in improving model performance.
(4) We release the code\footnote{\camerareadytext{https://github.com/junghwanjkim/multilingual\_aa}} and model\footnote{\camerareadytext{https://huggingface.co/Blablablab/multilingual-style-representation}} used in our experiments.

\section{Related Work}
\label{sec:related_work}
Recent studies~\cite{barlas2020crossdomain,fabien2020bertaa} have shown that Pre-trained Language Models (PLMs)~\cite{devlin2019bert,liu2019roberta} surpass traditional feature-based methods in authorship attribution.
AR learning~\cite{boenninghoff2019similarity,zhu2021idiosyncratic,riverasoto2021luar}, in particular, has emerged as a promising PLM-based approach, offering strong scalability to virtually unlimited numbers of authors.
AR methods use contrastive learning frameworks~\cite{oord2019infonce,khosla2020supconloss} to learn an embedding space that captures writing styles.
In this work, we investigate the multilingual generalization of AR learning.

\myparagraph{AR beyond English.}
While the majority of existing research on AR has focused on English datasets, there is a growing interest in extending AR methods to languages beyond English.
Earlier work in this direction primarily addresses individual low-resource languages, typically with datasets with at most a few hundred authors.
To overcome data scarcity, these studies either fine-tune the monolingual PLMs for the target language~\cite{avram2023bertbased,gabrovsek2023authorship,delanghe2024unsupervised,hossain2025authornet} or incorporate language-specific syntactic or morphological features~\cite{nitu2024authorship,misini2024automatic}.
However, the potential benefits of jointly training AR models across multiple languages remain unexplored, and the cross-lingual transfer capabilities of such models are largely unknown.
We fill this gap by proposing a multilingual framework, demonstrating that training AR models jointly on multiple languages leads to better authorship attribution accuracy across diverse linguistic contexts.


\myparagraph{Multilingual Semantic Representation.}
Our motivation stems from the success of multilingual semantic representation learning~\cite{artetxe2019massively,conneau2020xlmroberta}, which has shown strong cross-lingual transfer capability.
In this approach, fine-tuning a model on a source language improves performance on a target language~\cite{fujinuma2022match,philippy2023towards,chirkova2024key}.
However, it remains unclear whether AR models, which focus on capturing stylistic features of authorship rather than semantics, can enjoy similar cross-lingual transfer.
This work provides evidence that AR models can indeed enjoy cross-lingual transfer from multilingual training, which is surprising given the syntactic and grammatical diversity across languages.


\section{Proposed Method}
\label{sec:method}
We propose a multilingual AR method that trains a single model across multiple languages without relying on language-specific resources.
Our approach builds on supervised contrastive learning (\S\ref{subsec:method_scl}) and addresses shortcut issues in multilingual AR settings (\S\ref{subsec:method_shortcut}) through two key techniques.
First, \textit{Probabilistic Content Masking (PCM)} mitigates topic dependence by selectively masking content words (\S\ref{subsec:method_pcm}).
Second, \textit{Language-Aware Batching (LAB)} improves training efficiency and stability by promoting language-consistent batches, thereby reducing cross-lingual easy negatives and yielding stronger contrastive signals (\S\ref{subsec:method_lab}).

\subsection{Supervised Contrastive Learning}
\label{subsec:method_scl}
Our method adopts a supervised contrastive learning framework~\cite{khosla2020supconloss}, commonly used in recent state-of-the-art AR models~\citep[e.g.,][]{riverasoto2021luar,sawatphol2022topicreg}.
This framework promotes similarity between document pairs from the same author relative to similarity between pairs written by different authors.

Concretely, we train an AR model that maps input text $x$ to a vector representation $\mathbf{z}$.
Given a set of $N$ randomly sampled authors, we select two documents per author to form a document batch $B = \{ x_i^0, x_i^1 \}_{i \in [N]}$.
The contrastive loss for this batch is defined as
\begin{equation}
\label{eq:contrastive}
    \mathcal{L} = - \frac{1}{2N} \sum_{\substack{i \in [N] \\ k=0,1}} \log \frac{\exp \left(\mathbf{z}_i^k \cdot \mathbf{z}_i^{1-k} / \tau \right)}{ \sum_{\substack{j \in [N] \setminus \{ i \} \\ l=0,1}} \exp \left(\mathbf{z}_i^k \cdot \mathbf{z}_j^l / \tau \right)},
\end{equation}
where $\mathbf{z}_a^b$ denotes the representation of input $x_a^b$, and the dot product $\cdot$ denotes cosine similarity.
The temperature parameter $\tau$ controls the sharpness of the softmax distribution.
In each summand, $x_i^k$ is treated as the anchor and $x_i^{1-k}$ as the positive sample, and all $x_j^l$ for $j \neq i$ serve as negatives.

\subsection{Shortcut Learning in AR}
\label{subsec:method_shortcut}
Contrastive learning models are prone to shortcut learning~\cite{robinson2021shortcut,xue2023simplicitybias}, where they rely on easily accessible signals that are only spuriously correlated with the target task.
In multilingual AR, two prominent shortcuts are topic dependence and cross-lingual easy negatives.

\myparagraph{Topic Dependence.}
AR models may overfit to superficial topic shortcuts rather than capturing genuine stylistic features, impairing their ability to generalize across domains with different topic distributions~\cite{mikros2007topicinfluence}.
This issue arises because authors tend to write about recurring themes, which introduces topic biases into their documents~\cite{altakrori2021topicconfusion}.

Existing solutions follow two main strategies:
(1) controlling for topic using semantic information, and (2) removing topic-related words from the input.
The first strategy modifies training objectives~\cite{sawatphol2022topicreg,hu2024contrastdistaa} or batch construction~\cite{wegmann2022cav,fincke2024separating}, typically relying on topic models or semantic embeddings.
The second strategy removes topic-related words using part-of-speech taggers~\cite{wang2023luar} or topic models~\cite{man2024counterfactual}.
However, both approaches rely on language-specific tools, limiting their use in low-resource languages---a constraint we explicitly aim to eliminate.

\myparagraph{Cross-Lingual Easy Negatives.}
In multilingual AR settings, models can easily detect language mismatch between documents and assign low similarity to such cross-lingual pairs.
When negative pairs come from different languages, their already low similarity leads to weak contrastive signals.
Because their similarity to the anchor is already low, these examples yield negligible gradient updates and offer little training signal.
This weak supervision not only slows convergence but can also cause numerical instability when the denominator in Equation~\ref{eq:contrastive} approaches zero.

\subsection{Probabilistic Content Masking (PCM)}
\label{subsec:method_pcm}
PCM reduces \textit{topic dependence} in AR learning by randomly masking tokens that are less indicative of writing style.
During training, PCM randomly masks content words---those that carry meaning---thereby encouraging the model to focus on function words, which express grammatical structure.
Because authors tend to use function words more consistently than content words across documents~\cite{argamon2005usefulnessfunctionwords,kestemont2014functionwordstheory}, this shift in focus promotes the learning of stylistic, rather than topical, patterns.

To identify content words without relying on language-specific tools, we adopt an approximate strategy based on subword token frequency.
While conventional approaches use predefined stopword lists or part-of-speech taggers to distinguish content from function words~\cite{zhu2021idiosyncratic,wang2023luar}, such tools are often unavailable or unreliable for low-resource languages.
Moreover, while multilingual neural taggers exist, these introduce substantial computational overhead, making them prohibitively expensive for training with large corpora like those used here.
Instead, we treat high-frequency subword tokens from the training corpus as function tokens, eliminating dependence on external tools.
This frequency-based approach is lightweight and adapts naturally to the token distribution of each domain, which is particularly effective when prevalent function words differ across languages and domains.
\camerareadytext{In English, a few methods have used word-frequency heuristics to always remove content words as a preprocessing step for authorship attribution~\citep{stamatatos2017distortion,markov2018highlyfrequentwords}.}

Randomness in PCM plays a key role in both regularizing training and preserving stylistic cues in content tokens.
By varying the masked tokens across training steps, PCM exposes the model to different views of the same input, effectively serving as data augmentation.
In addition, random masking allows some content tokens to remain unmasked, enabling the model to capture consistent lexical choices that may reflect an author's stylistic signature.
Together, these effects enhance the model's ability to learn effective ARs.

\subsection{Language-Aware Batching (LAB)}
\label{subsec:method_lab}
LAB mitigates the \textit{cross-lingual easy negative} problem that arises in multilingual AR settings.
Unlike standard batching based on random shuffling---which mixes documents regardless of language---LAB constructs batches by shuffling training data within each language, ensuring that each batch contains documents in the same language.
The order of languages is also permuted each epoch to reduce potential training bias.
This batching strategy improves both efficiency and stability by ensuring greater language consistency within batches, thereby reducing low-signal cross-lingual negatives and facilitating more effective contrastive learning.


\section{Experimental Setup}
\label{sec:setup}
We evaluate the effectiveness and generalizability of our multilingual AR model, which is jointly trained across multiple languages, in two settings.
First, we compare its performance to monolingual models trained on the corresponding target language to assess the effectiveness of multilingual training (\S\ref{subsec:results_vs_monolingual}).
Next, to evaluate cross-lingual and cross-domain generalization, we compare our model to a monolingual English model on languages and domains unseen during training by either model (\S\ref{subsec:results_vs_english}).

\myparagraph{Dataset.}
Our dataset comprises over 6.2M authors across 59 languages and 17 diverse domains, including online forums (Reddit), product reviews (Amazon), novels (BookCorpus), and academic articles (PubMed).
\camerareadytext{Appendix Table~\ref{tab:setup_desc_english} provides the descriptions of these domains in our dataset.}
All non-English documents are sourced from four Wikipedia domains---articles, user pages, talk pages, and user talk pages\footnote{User pages are for interpersonal discussion, notices, testing, and personal content. Talk and user talk pages are for discussion of article and user page improvements, respectively.}---which collectively represent a broad range of discourse types.
\camerareadytext{The languages span 20 language families and 19 distinct script systems.
Appendix Tables~\ref{tab:setup_langfamily_multi_seen} and \ref{tab:setup_langfamily_multi_unseen} show the language family and the script system for each language.}

The \textit{seen} subset, used for training, includes 35 languages and 10 domains.
In Section~\ref{subsec:results_vs_monolingual}, the comparison between our multilingual model and monolingual baselines focuses on the 22 seen languages that contain a sufficient number of authors to support effective monolingual AR training.
The remaining 24 languages and 7 domains constitute the \textit{unseen} subset, which is reserved for evaluating cross-lingual and cross-domain generalization in Section~\ref{subsec:results_vs_english}.
The seen data is split into training, validation, and test sets using an 85/5/10 split;
the unseen data is split into validation and test sets with a 33.3/66.7 ratio.
Models are trained on the training set, hyperparameters are tuned on the validation set, and results are reported on the test set.
\camerareadytext{Appendix Tables~\ref{tab:setup_stats_english}, \ref{tab:setup_stats_multi_seen}, and \ref{tab:setup_stats_multi_unseen} present the dataset statistics for English domains, multilingual seen languages, and unseen languages, respectively.}

Each author is associated with exactly two documents written in the same language.
During training, these two documents form a positive contrastive pair.
For evaluation, one document is used as a \textit{query} (an author to be found) and their other document is denoted as the \textit{candidate}.

\myparagraph{Metrics.}
Following the standard practice in the literature~\cite{riverasoto2021luar,sawatphol2022topicreg,man2024counterfactual,fincke2024separating}, we evaluate AR models using an authorship attribution task and report Recall@8 (R@8) and Mean Reciprocal Rank (MRR).
For each query document, all candidate documents are ranked by cosine similarity in the AR embedding space.
R@8 measures the portion of queries for which the correct author appears among the top 8 candidates, while MRR averages the reciprocal rank of the correct candidates over all queries.\footnote{\camerareadytext{These definitions differ slightly from the standard ones, but remain valid in our setting since each query has exactly one corresponding candidate.}}



\myparagraph{Training Details.}
Our multilingual AR model is fine-tuned from XLM-RoBERTa-large\footnote{\camerareadytext{https://huggingface.co/FacebookAI/xlm-roberta-large}}~\cite{conneau2020xlmroberta}, a multilingual Transformer pretrained on 100 languages.\footnote{\camerareadytext{We also train and evaluate Llama3.2-1B (https://huggingface.co/meta-llama/Llama-3.2-1B); however, due to its more limited language coverage, it underperforms compared to XLM-RoBERTa-large on low-resource languages. As a result, we report its results in the appendix. Additional training details for Llama model can be found in Appendix~\ref{subsec:appendix_setup_train}.}}
We train our AR model for 5 epochs with a learning rate of 1e-4, batch size of 1{,}024, and a masking rate of 0.2.
The learning rate and batch size are chosen based on our preliminary experiments, while the masking rate is determined via hyperparameter search,  discussed later in Section~\ref{subsec:analysis_maskrate}.
\camerareadytext{We use the AdamW optimizer~\cite{loshchilov2018decoupled} and the WSD learning rate schedule~\cite{hu2024minicpm}.
The temperature parameter in the contrastive loss is set to $\tau = 0.1$.}
We select the best checkpoint based on validation loss and use it as our final model.
\camerareadytext{All models are implemented with PyTorch-Lightning and the Huggingface Transformer library.}

\myparagraph{Baselines.}
For comparison, we train monolingual AR models for each target language.
Each model is fine-tuned from XLM-RoBERTa-large---the same base model used for our multilingual model---using only data from the corresponding language.
Due to the smaller size of each monolingual dataset, we train for more epochs (ranging from 20 to 50) until convergence and select the best checkpoint based on validation loss.
We refer to these models as \textit{monolingual XLM-RoBERTa} models.
When the target language is English, we refer to the model as the \textit{English-only XLM-RoBERTa} model.
Additionally, for 8 languages with well-established monolingual BERT-like models (Appendix Table~\ref{tab:exp1_mono_berts}), we fine-tune those using the same training setup and refer to the resulting models as \textit{monolingual BERT} models.
When unspecified, the base model defaults to the XLM-RoBERTa-large model.

We also evaluate against a state-of-the-art multilingual AR method, mStyleDistance~\cite{qiu2025mstyledistance}, but only show these results in Appendix Tables~\ref{tab:exp1_full_r@8} and \ref{tab:exp1_full_mrr} for visual clarity when plotting, as it performs poorly in our setup.

\begin{figure}[t]
    \centering
    \includegraphics[width=\linewidth]{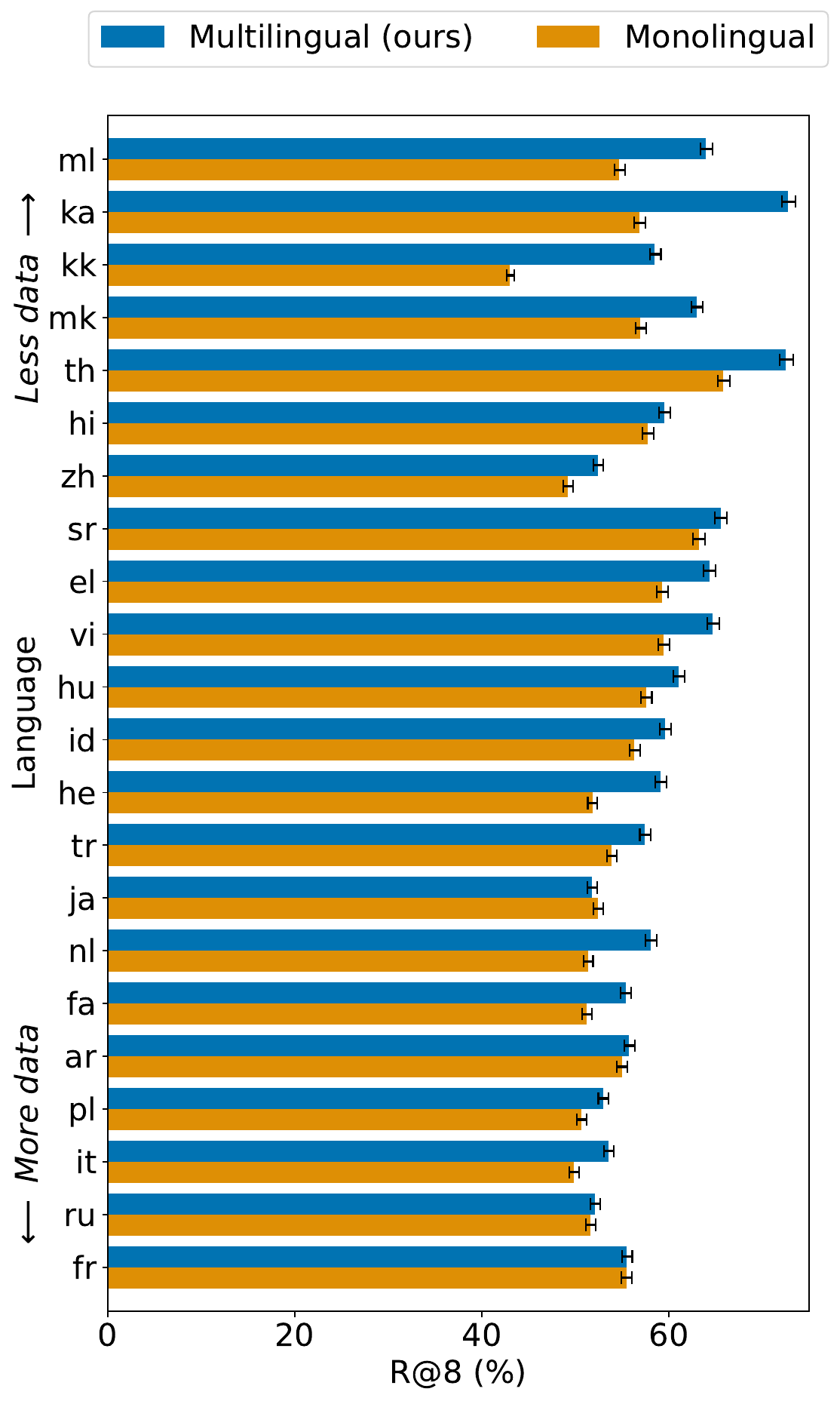}
    \caption{\textbf{Multilingual training provides consistent gains across languages.}
    A single \textit{multilingual} model trained across multiple languages outperforms 21 out of 22 \textit{monolingual} models, each trained on its respective language.
    Languages with less data show greater gain. For all plots, error bars show 95\% confidence intervals.}
    \label{fig:exp1_across_language_xlmroberta}
\end{figure}

\section{Experimental Results}
\label{sec:results}
This section presents the performance of our multilingual models across languages and domains, following the evaluation setup described in Section~\ref{sec:setup}.

\subsection{Effectiveness across Languages}
\label{subsec:results_vs_monolingual}

Multilingual training consistently improves authorship attribution across languages.
Incorporating training data from multiple languages enables the model to outperform monolingual models trained solely on language-specific data.
Our multilingual model outperforms monolingual baselines on all evaluated languages except Japanese, with an average R@8 improvement of 4.85\% across languages (Figure~\ref{fig:exp1_across_language_xlmroberta}).
This result indicates that our multilingual model captures stylistic features that transfer across languages---a notable finding given the diversity of 22 evaluated languages, which span 15 language families and 11 script systems.

\begin{figure}[t]
    \centering
    \includegraphics[width=0.99\linewidth]{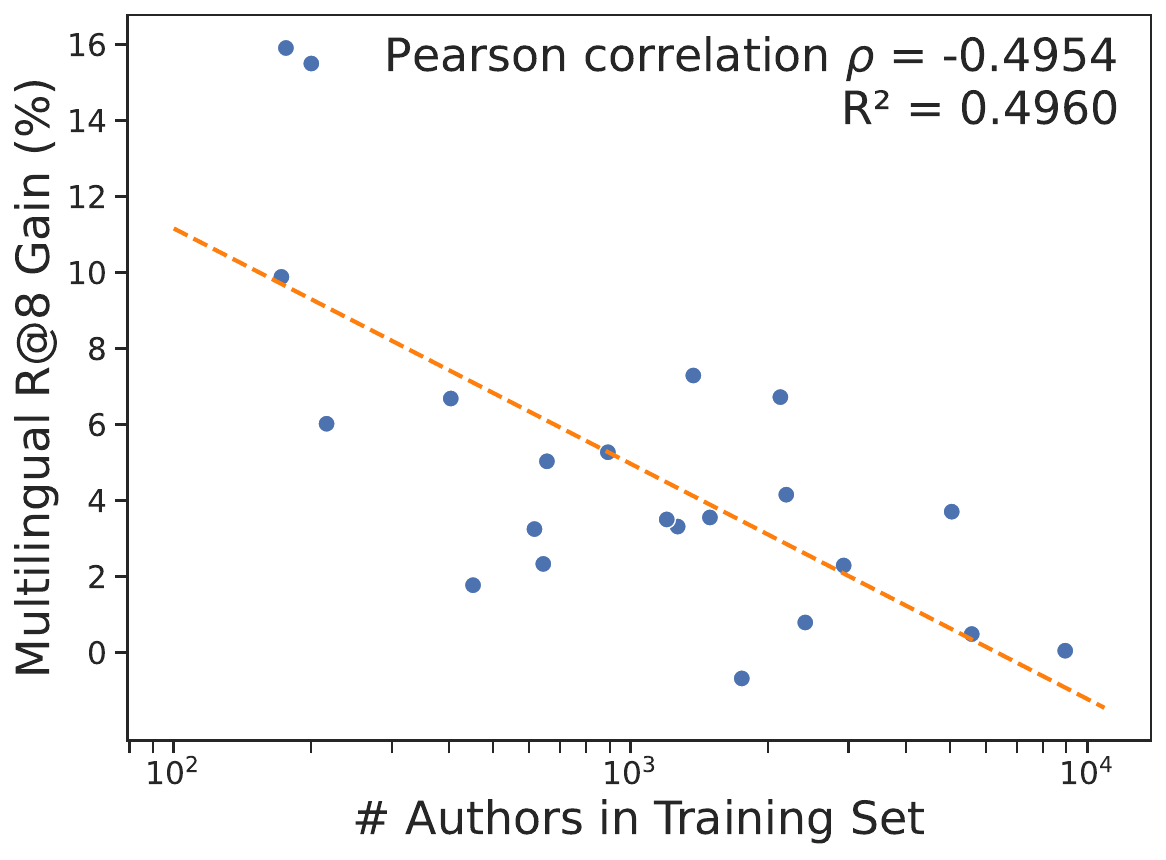}
    \caption{\textbf{Multilingual training yields greater gains for low-resource languages.}
    There is a strong negative correlation between the number of training authors and the R@8 improvement from multilingual training.}
    \label{fig:exp1_gain_by_datasize}
\end{figure}

Low-resource languages benefit more substantially from multilingual training.
For example, R@8 improves by over 15\% for languages with less data, such as Kazakh and Georgian.
To investigate this trend, we plot R@8 gains from multilingual training against the number of authors in each language dataset (Figure~\ref{fig:exp1_gain_by_datasize}).
The results reveal a clear negative correlation: as the number of authors decreases, the benefits of multilingual training increase.
This pattern demonstrates that multilingual training enables cross-lingual transfer, effectively redistributing representation capacity from high-resource to low-resource languages, and making this approach particularly effective for underrepresented languages.

Our multilingual model even outperforms monolingual BERT models\footnote{\camerareadytext{Appendix Table~\ref{tab:exp1_mono_berts} lists the monolingual base models that we used to train monolingual BERT AR models.}} fine-tuned from language-specific base models.
These monolingual BERT models serve as stronger baselines than monolingual XLM-RoBERTa models, as their tokenizers and pretraining corpora are tailored to the target language.
Nonetheless, our multilingual model surpasses them in 6 out of 8 evaluated languages---excluding only French and Polish---despite lacking any language-specific tokenization (Table~\ref{tab:exp1_full_r@8}).
This further highlights the effectiveness of our multilingual approach.



\subsection{Cross-Lingual and Cross-Domain Generalization}
\label{subsec:results_vs_english}

\begin{figure}[t]
    \centering
        \centering
        \includegraphics[width=0.99\columnwidth]{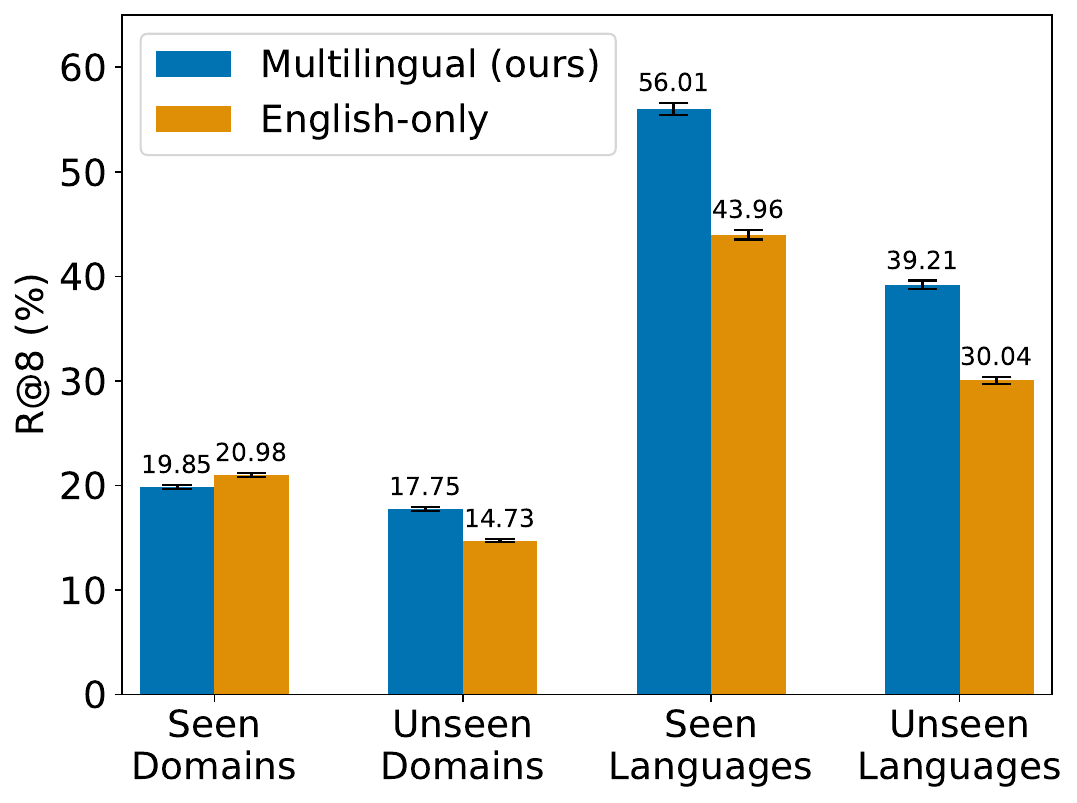}
    \caption{\textbf{The multilingual model exhibits stronger cross-lingual and cross-domain generalization than its English-only counterpart.}
    Multilingual training improves R@8 on unseen English domains, seen languages, and unseen languages, while incurring only a minimal drop in the seen English domain.}
    \label{fig:exp2_cross_lingual_domain_xlmroberta}
\end{figure}

Multilingual training substantially improves generalization to unseen languages.
Our multilingual model outperforms its English-only counterpart by 9.17\% in R@8, a relative improvement of 30.52\% (Figure~\ref{fig:exp2_cross_lingual_domain_xlmroberta}).
Notably, these gains extend to Armenian---a language with no shared script with the languages seen in training---and Telugu---a language with no shared language family  in our training data (Table~\ref{tab:full_multi_r@8}).
These results show that multilingual training enables generalization well across scripts and language families.

Multilingual training also enhances generalization to unseen English domains.
Our model achieves a 3.02\% gain in R@8 over the English-only model, indicating that exposure to diverse languages encourages the learning of stylistic features that are domain-agnostic.
This, in turn, improves robustness to novel domains, making our multilingual models more suitable for general-purpose writing style representation.

The multilingual model performs strongly on seen languages, consistent with the improvements observed for individual languages in Section~\ref{subsec:results_vs_english}.
Our model outperforms the English-only baseline on these languages, increasing R@8 from 43.96\% to 56.01\%, a relative improvement of 27.41\%.

These gains come at a minimal cost on English seen domains, where the multilingual model lags behind the English-only baseline by only 1\% in R@8.
This minor decline, combined with the substantial improvements in cross-lingual and cross-domain settings, supports the hypothesis that multilingual training promotes the acquisition of universal stylistic representations in authorship modeling.
As generalization outweighs in-domain accuracy in general-purpose authorship modeling, the multilingual model provides a more suitable solution than the English-only model.


\section{Performance and Ablation Analyses}
\label{sec:analysis}
This section presents five additional analyses. The first three assess aspects of our model:  a hyperparameter search to examine sensitivity to masking rate (\S\ref{subsec:analysis_maskrate}) and ablation studies to assess the contribution of PCM (\S\ref{subsec:analysis_ablation_pcm}) and LAB (\S\ref{subsec:analysis_ablation_lab}). The remaining two demonstrate its generalization to new settings through evaluation on two downstream authorship tasks: authorship verification (\S\ref{subsec:analysis_authorship_verification}) and machine-generated text detection (\S\ref{subsec:analysis_ai_detection}).

\begin{figure}[t]
    \centering
    \includegraphics[width=0.99\columnwidth]{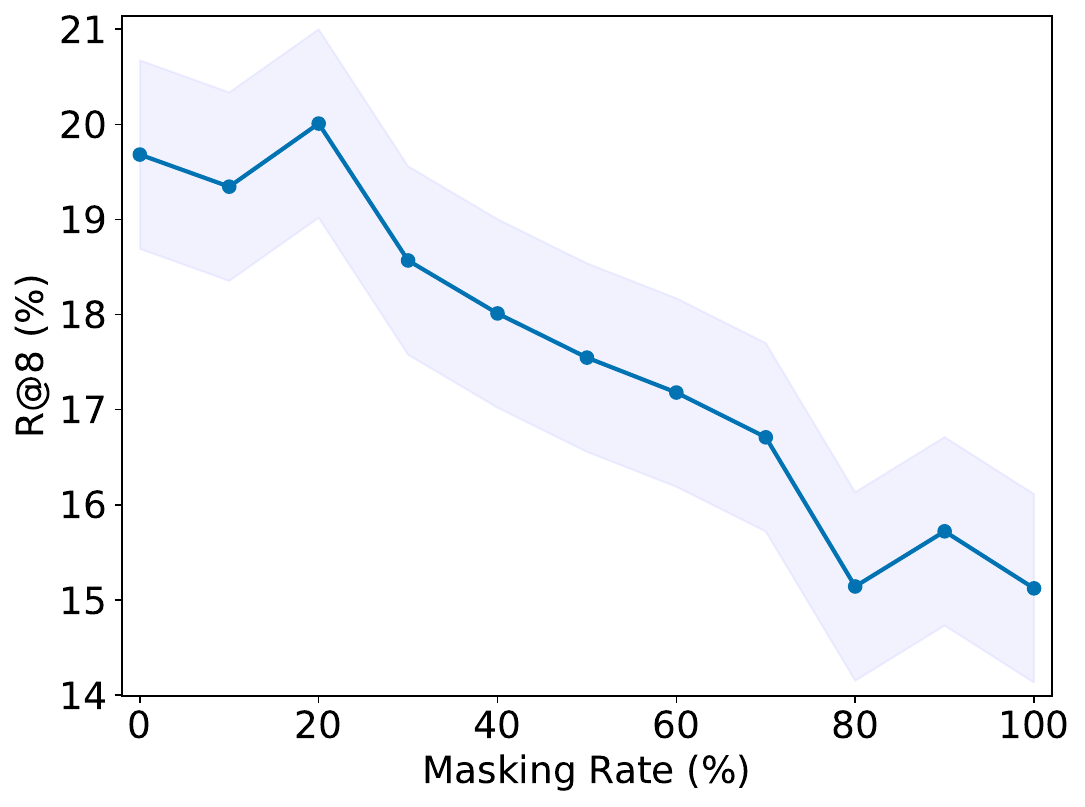}
    \caption{\textbf{Hyperparameter search identifies 20\% as the optimal masking rate.}
    We search using a 10\% random subset of the training data and report R@8 on the validation set for seen English domains.}
    \label{fig:analysis1_maskrate_xlmroberta}
\end{figure}

\subsection{Masking Rate Search}
\label{subsec:analysis_maskrate}

We selected the masking rate in earlier experiments via hyperparameter search by training models on a 10\% random subset of the training set, evaluating R@8 on the validation set for seen English domains. A 20\% masking rate yielded the best performance (Figure~\ref{fig:analysis1_maskrate_xlmroberta}) and was used for full-scale training.

\subsection{Ablation Study: PCM}
\label{subsec:analysis_ablation_pcm}
This section evaluates the effectiveness of our proposed PCM for authorship representation.

\myparagraph{Setup.}
We compare PCM against four state-of-the-art AR methods and two baseline masking approaches.
We restrict evaluation to the English portion of the dataset, as most AR methods are designed specifically for English, and masking baselines depend on language-specific resources not readily available in other languages.
For state-of-the-art AR baselines, we include LUAR~\cite{riverasoto2021luar}, CAV~\cite{wegmann2022cav}, StyleDistance~\cite{patel2025styledistance}, and mStyleDistance~\cite{qiu2025mstyledistance}.
For masking baselines, we evaluate Stopword~\cite{zhu2021idiosyncratic}, which retains only tokens from the predefined SpaCy stopword list, and POS~\cite{wang2023luar}, which masks proper nouns identified using the Stanza part-of-speech tagger.
Additionally, we include a random ordering baseline as a lower-bound reference to contextualize the results.

\begin{table}[t]
\centering
\resizebox{\columnwidth}{!}{
    \begin{tabular}{llrr}
    \textbf{Model} & \textbf{Mask} & \textbf{Seen} & \textbf{Unseen} \\ 
    \hline
    RandomOrder & - & 0.0016  & 0.0023  \\
    \hline
    LUAR & \xmark & 10.35  & 11.79  \\
    CAV & \xmark & 1.50  & 2.75  \\
    StyleDistance & \xmark & 3.16  & 4.86  \\
    \hline
    \multirow{4}{*}{RoBERTa}
    & \xmark & 22.60  & 14.64  \\
    & Stopword & 10.34  & 7.88  \\
    & POS & 19.14  & 12.97  \\
    & PCM & \textbf{24.66} & \textbf{16.24} \\
    \end{tabular}
}
\caption{\textbf{PCM achieves the highest R@8 among all evaluated methods on both seen and unseen English domains.}
It outperforms state-of-the-art AR methods by a substantial margin and surpasses masking baselines that rely on language-specific resources.
For all tables, the best performance is bolded.}
\label{tab:analysis2_pcm}
\end{table}

\myparagraph{Results.}
PCM achieves the best performance among all AR methods on both seen and unseen English domains (Table~\ref{tab:analysis2_pcm}).
Among the AR baselines, LUAR performs best, likely due to its use of a contrastive training objective---similar to ours---that aligns well with the authorship attribution task.
In contrast, CAV and StyleDistance perform worse, as they are not optimized for authorship attribution but instead focus on reducing topic dependence.
Still, all AR baselines achieve non-trivial R@8 scores and substantially outperform the random baselines.

PCM consistently outperforms all masking baselines on both domain splits.
Surprisingly, Stopword and POS masking baselines underperform even the no-masking baseline, despite relying on high-quality, language-specific resources to identify function words.
This underperformance is likely due to their deterministic masking schemes, which create a mismatch between masked training inputs and unmasked test inputs, hindering the model's ability to learn consistent stylistic signals.
We hypothesize that stochastic variants of these approaches could improve performance on English data; however, we leave this exploration to future work, as these methods are not applicable to our multilingual training setup, which is the primary focus of this study.

\begin{table}[t]
\centering
    \begin{tabular}{lrrrr}
    \multirow{2}{*}{\textbf{LAB}} & \multicolumn{2}{c}{\textbf{English}} & \multicolumn{2}{c}{\textbf{Multilingual}} \\ 
    \cmidrule{2-5}
    & \textbf{Seen} & \textbf{Unseen} & \textbf{Seen} & \textbf{Unseen} \\ 
    \hline
    \xmark & \textbf{20.52} & 15.49 & 49.82 & 32.82 \\
    \cmark & 19.85 & \textbf{17.75} & \textbf{56.01} & \textbf{39.21} \\
    \end{tabular}
\caption{\textbf{LAB improves multilingual R@8 and enhances both cross-lingual and cross-domain generalization in terms of R@8.}
These gains in unseen English domains and multilingual subsets come at only a minor cost in seen English domains.}
\label{tab:analysis3_lab}
\end{table}

\subsection{Ablation Study: LAB}
\label{subsec:analysis_ablation_lab}
This section ablates LAB to see its impact on performance across languages and domains.

\myparagraph{Setup.}
We train the multilingual AR model with and without LAB and compare their performance.
With LAB, each training batch consists primarily of documents in the same language.
Without LAB, batches are formed via random shuffling, mixing documents from different languages.
All other training details remain identical to those used in the main experiments.

\myparagraph{Results.}
LAB consistently outperforms the random batching baseline, achieving over 2\% higher R@8 in unseen English domains and more than 6\% gains in both seen and unseen languages, while incurring less than a 1\% drop in seen English domains (Table~\ref{tab:analysis3_lab}).
These results show that LAB improves cross-lingual and cross-domain generalization by reducing easy negatives during training.


\subsection{Authorship Verification}
\label{subsec:analysis_authorship_verification}
We evaluate the zero-shot downstream performance of our multilingual AR model in two authorship verification settings that differ from authorship attribution used in previous sections.

\myparagraph{Setup.}
We use the authorship verification task from the PAN shared tasks, where the goal is to determine whether a given pair of documents was written by the same author.
Evaluation is conducted on the test corpora from the 2013 to 2015 PAN datasets~\cite{juola2013pan,stamatatos2014pan,stamatatos2015pan}, covering Greek, Spanish, and Dutch.
The training dataset for our multilingual AR model includes Greek and Dutch, but not Spanish.
For zero-shot prediction, we compute cosine similarity between AR embeddings and use it as the prediction score, reporting AUROC as the evaluation metric.
We compare against the same state-of-the-art AR baselines used in Section~\ref{subsec:analysis_ablation_pcm}: LUAR, CAV, StyleDistance, and mStyleDistance.

\begin{figure}[t]
    \centering
        \centering
        \includegraphics[width=\columnwidth]{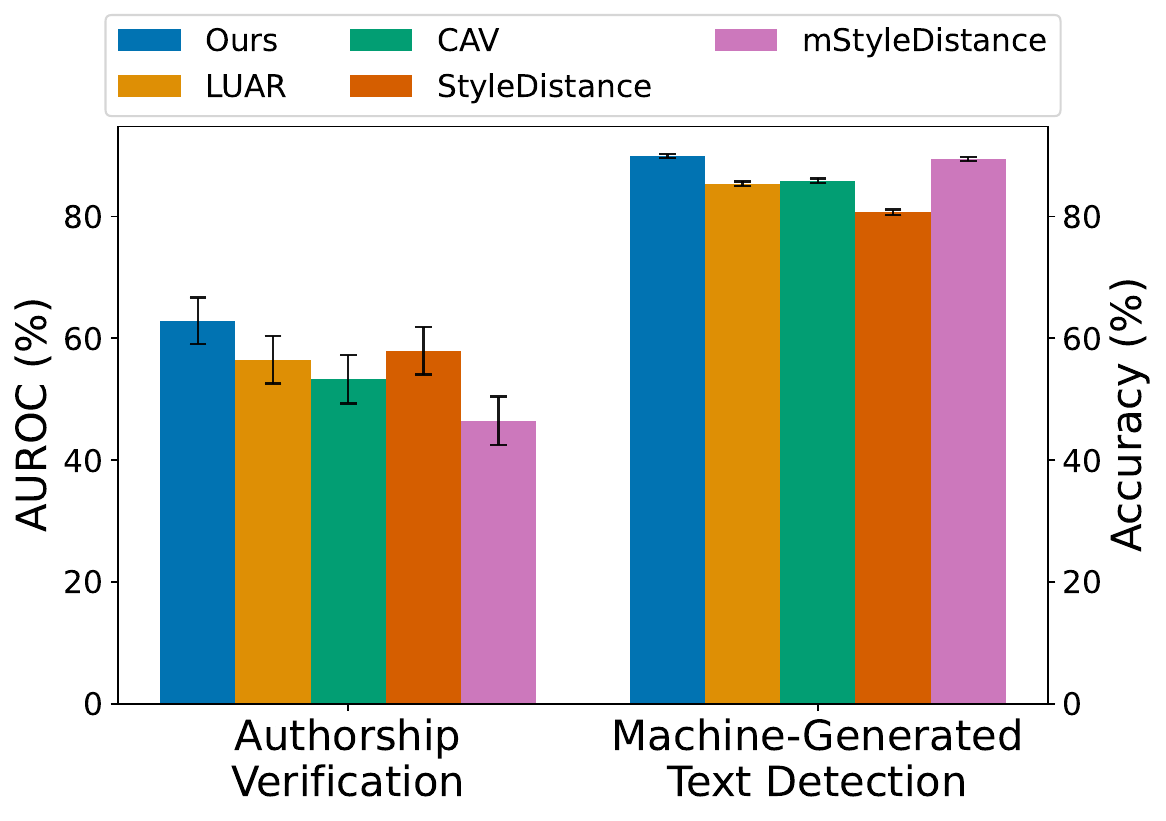}
    \caption{\textbf{The multilingual models show the best overall performance in authorship verification tasks.} Multilingual training generalizes competitively to out-of-domain settings.}
    \label{fig:analysis45_downstream}
\end{figure}

\myparagraph{Results.}
Our model outperforms all evaluated AR baselines (Figure~\ref{fig:analysis45_downstream}, left).
Across the 8 splits, our model achieves the highest AUROC in 3 and the second-highest in another 3 (Table~\ref{tab:analysis4_pan_av_full}).
These results highlight the strong generalization ability of our model to out-of-domain and task-shifted authorship scenarios.

\subsection{Machine-Generated Text Detection}
\label{subsec:analysis_ai_detection}
We examine the effectiveness of our multilingual AR model on machine-generated text detection.

\myparagraph{Setup.}
Machine-generated text detection requires distinguishing between human-written and machine-generated texts.
We use the MULTITuDE dataset~\cite{macko2023multitude}, which includes 11 languages: English, Spanish, Russian, Dutch, Catalan, Czech, German, Chinese, Portuguese, Arabic, and Ukrainian.
The training dataset for our AR model covers a subset of these languages---specifically, English, Russian, Dutch, Chinese, and Arabic---while the remaining 6 are unseen during training.
We apply logistic regression with default parameters using scikit-learn~\cite{pedregosa2011scikit-learn} on top of the AR embeddings and report prediction accuracy.
We compare against the same baselines as in the previous experiment. 

\myparagraph{Results.}
Our multilingual AR model achieves the strongest overall accuracy among all evaluated methods (Figure~\ref{fig:analysis45_downstream}, right).
Our model ranks first in 5 languages and second in another 5, demonstrating consistent effectiveness across both seen and unseen languages (Table~\ref{tab:analysis5_multitude_ai_full}). This result demonstrates our techniques enable representations that generalize well even to machine writing.


\section{Conclusion}
\label{sec:conclusion}
In this paper, we introduce a novel method for multilingual authorship representation that integrates two core techniques:
(1) PCM, which selectively masks content words to drive the model's focus toward stylistic cues; and
(2) LAB, which groups training data by language to avoid cross-lingual easy negatives and improve contrastive learning efficiency.
These techniques address key challenges in authorship modeling, including data scarcity and topic dependence.
Our experiments show that the proposed multilingual model consistently outperforms monolingual baselines, with particularly strong gains in low-resource languages.
Furthermore, our method improves performance in previously unseen languages and domains.
Further analysis examines the effect of the masking rate, evaluates the contributions of PCM and LAB, and demonstrates the effectiveness of our model on downstream tasks.
This work demonstrates that multilingual training can significantly enhance authorship representation models, opening new possibilities for multilingual authorship analysis.
\camerareadytext{Future directions for this work include: (1) extending LAB to enhance cross-domain generalization; (2) incorporating a broader range of domains into the multilingual author-labeled dataset to improve its diversity and robustness; and (3) exploring the application of authorship representations to other style-related tasks.}

\section*{Limitations}
We limit our settings to the case where each author only writes in one language.
While our multilingual models allow for cross-language AR, whether accurate authorship attribution for authors writing in multiple languages is possible with our model remains unclear.
Cross-lingual authorship attribution differs fundamentally from our current focus and is left for future work.

While we have sourced English language data from multiple diverse domains, our multilingual dataset is sourced from a single domain: Wikipedia.
AR dataset remains rare, and Wikipedia is the largest data source that is feasible to collect for a high number of languages to ensure linguistic diversity.
While Wikipedia contains multiple genres of text (e.g., articles, talk pages), these do not reflect the diversity of our English data.
\camerareadytext{Moreover, the article pages for low-resource languages often exhibit quality issues~\cite{tatariya2025wikipediaquality}, including duplicate entries, bot-generated content, and the presence of foreign scripts.}
As a result, our evaluation likely does not reflect the full cross-domain generalization capability in non-English languages.

\section*{Ethical Considerations}
In this work, we present a framework for developing a multilingual AR model.
AR is a dual-use technology, and models have been used to both reveal and hide the identity of an anonymous author using stylistic difference measurement and style replacement, respectively.
Positive applications of the technology have been used in applications like historical document attribution \citep[e.g.,][]{gurney1998authorship,juola2008authorship} and for identifying likely authors of documents related to criminal activity \citep[e.g.,][]{olsson2009wordcrime,saxena2023idtraffickers}. However, negative applications could also be used to remove anonymity from individuals in sensitive situations (e.g,. where their demographics or some aspect of their identity puts them at risk of cultural retribution) or where political agents look to pursue individuals. 

\section*{Acknowledgments}

\camerareadytext{This research is supported in part by the Office of the Director of National Intelligence (ODNI), Intelligence Advanced Research Projects Activity (IARPA), via the HIATUS Program contract \#2022-22072200006. The views and conclusions contained herein are those of the authors and should not be interpreted as necessarily representing the official policies, either expressed or implied, of ODNI, IARPA, or the U.S. Government. The U.S. Government is authorized to reproduce and distribute reprints for governmental purposes notwithstanding any copyright annotation therein.}


\bibliography{reference}

\newpage
\appendix

\setcounter{figure}{0}
\setcounter{table}{0}
\renewcommand*\thefigure{A\arabic{figure}}
\renewcommand*\thetable{A\arabic{table}}

\clearpage
\section{Supplementary Material for Section~\ref{sec:setup}}
\label{sec:appendix_setup}
This section presents supplementary material for the experimental setup.

\subsection{Training Details for Llama model.}
\label{subsec:appendix_setup_train}
Since Llama3.2 is a causal language model, we modify its attention layers to enable bidirectional context, following LLM2Vec~\cite{behnamghader2024llm2vec}, and pool over tokens in the final layer representations.
We also apply LoRA~\cite{hu2022lora} to Llama3.2 to enable training with larger batch sizes, which we find essential for effective AR learning.

\begin{table}[t]
\centering
\resizebox{\columnwidth}{!}{
    \begin{tabular}{ll}
    \textbf{Language} & \textbf{Model} \\
    \hline
    French  & almanach/camembert-large \\
    Italian & Musixmatch/umberto-commoncrawl-cased-v1 \\
    Polish  & sdadas/polish-roberta-large-v2 \\
    Farsi   & HooshvareLab/roberta-fa-zwnj-base \\
    Dutch   & DTAI-KULeuven/robbert-2023-dutch-large \\
    Hebrew  & HeNLP/HeRo \\
    Indonesian & flax-community/indonesian-roberta-large \\
    Hungarian & uvegesistvan/Hun\_RoBERTa\_large\_Plain \\
    \end{tabular}
}
\caption{Monolingual base models used to train our monolingual BERT AR models.}
\label{tab:exp1_mono_berts}
\end{table}

\section{Supplementary Material for Section~\ref{sec:results}}
\label{sec:appendix_results}
We provide additional results of our main experiments.

\subsection{Effectiveness across Languages}

The full list of R@8 and MRR scores is shown in Table~\ref{tab:exp1_full_r@8} and Table~\ref{tab:exp1_full_mrr}, respectively.

The effectiveness of multilingual AR modeling depends not only on model size but also on the degree of multilingual support in the underlying encoder.
Although Llama model is more than twice the size of XLM-RoBERTa, it outperforms XLM-RoBERTa in only 10 out of 22 languages, with an average gain of 2\%.
The languages where Llama performs better tend to be high-resource, reflecting its bias toward languages with large pre-training corpora.
In contrast, XLM-RoBERTa outperforms Llama in 12 languages, with gaps of up to 20\% in low-resource languages such as Hebrew, Georgian, and Malayalam.
Notably, XLM-RoBERTa performs better in Hindi and Thai---languages that Llama officially supports---while Llama outperforms XLM-RoBERTa in Russian and Polish, which are not officially supported by Llama.
These results underscore the importance of broad multilingual coverage in the pertaining phase of the language model for effective multilingual authorship representation.

The comparison between multilingual and monolingual variants also reveals important distinctions between the two model families.
For most languages, the monolingual Llama models outperform their multilingual counterpart, although the margin is smaller than that observed between monolingual and multilingual XLM-RoBERTa models.
However, this gap should be interpreted with caution.
Llama's vocabulary is not optimized for many low-resource languages, often resulting in subword tokenization that splits text into characters or even bytes.
For example, the number of tokens generated for the same document in Georgian or Malayalam is over five times higher in Llama than in XLM-RoBERTa, making it significantly more difficult for the model to capture universal stylistic signals.
Therefore, the performance gap between monolingual and multilingual LLaMA models is not a reliable indicator of the effect of multilingual training alone.

\begin{figure}[t]
    \centering
        \centering
        \includegraphics[width=\columnwidth]{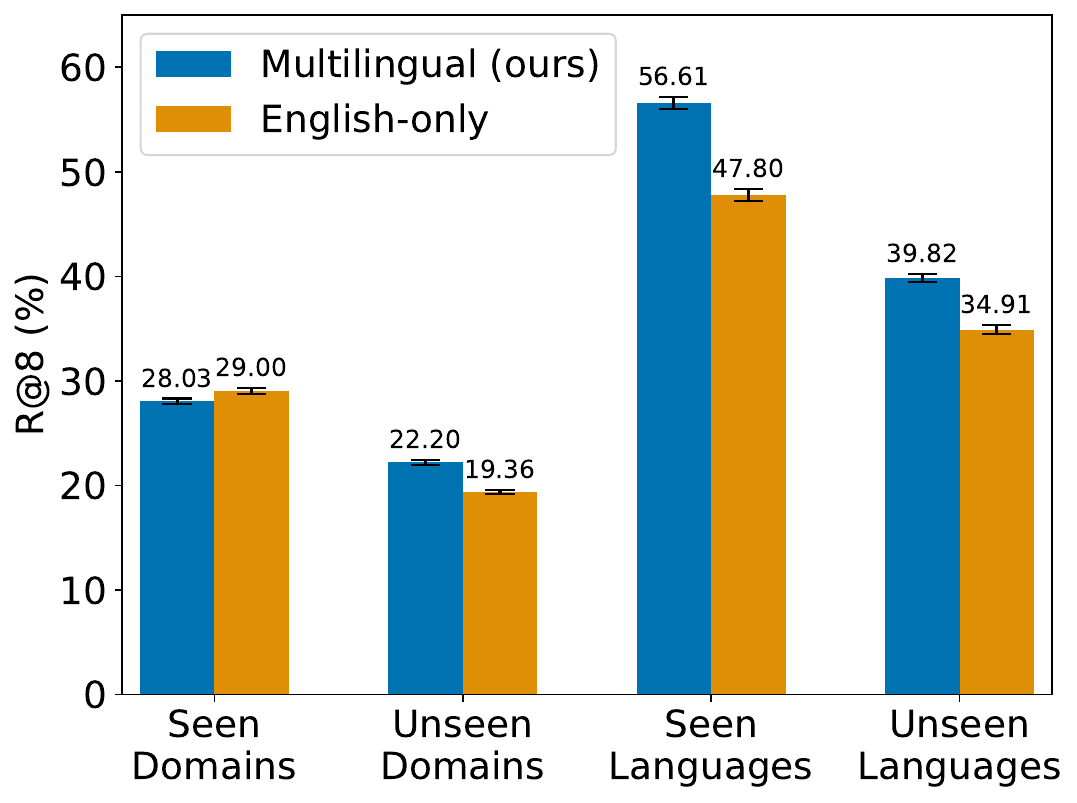}
    \caption{\textbf{The multilingual Llama model also exhibits stronger cross-lingual and cross-domain generalization than its English-only counterpart.}
    Multilingual training improves R@8 on unseen English domains, seen languages, and unseen languages, while incurring only a minimal drop in the seen English domain.}
    \label{fig:exp2_cross_lingual_domain_llama}
\end{figure}

\subsection{Cross-Lingual and Cross-Domain Generalization}
We repeat the experiments for Figure~\ref{fig:exp2_cross_lingual_domain_xlmroberta} with Llama3.2-1B, and the result is shown in Figure~\ref{fig:exp2_cross_lingual_domain_llama}.
As in the XLM-RoBERTa-large case, multilingual training improves cross-lingual and cross-domain generalization at a small cost in seen English domain performance.
Table~\ref{tab:exp2_cross_lingual_domain_mrr} presents the corresponding MRR results for both XLM-RoBERTa-large and Llama3.2-1B, which show a similar trend to R@8 results.

The full per-domain R@8 and MRR across all evaluated models are available in Table~\ref{tab:full_english_r@8} and Table~\ref{tab:full_english_mrr}, respectively.
The full per-language R@8 and MRR across all evaluated models are available in Table~\ref{tab:full_multi_r@8} and Table~\ref{tab:full_multi_mrr}, respectively.

\begin{figure}[t]
    \centering
    \includegraphics[width=\columnwidth]{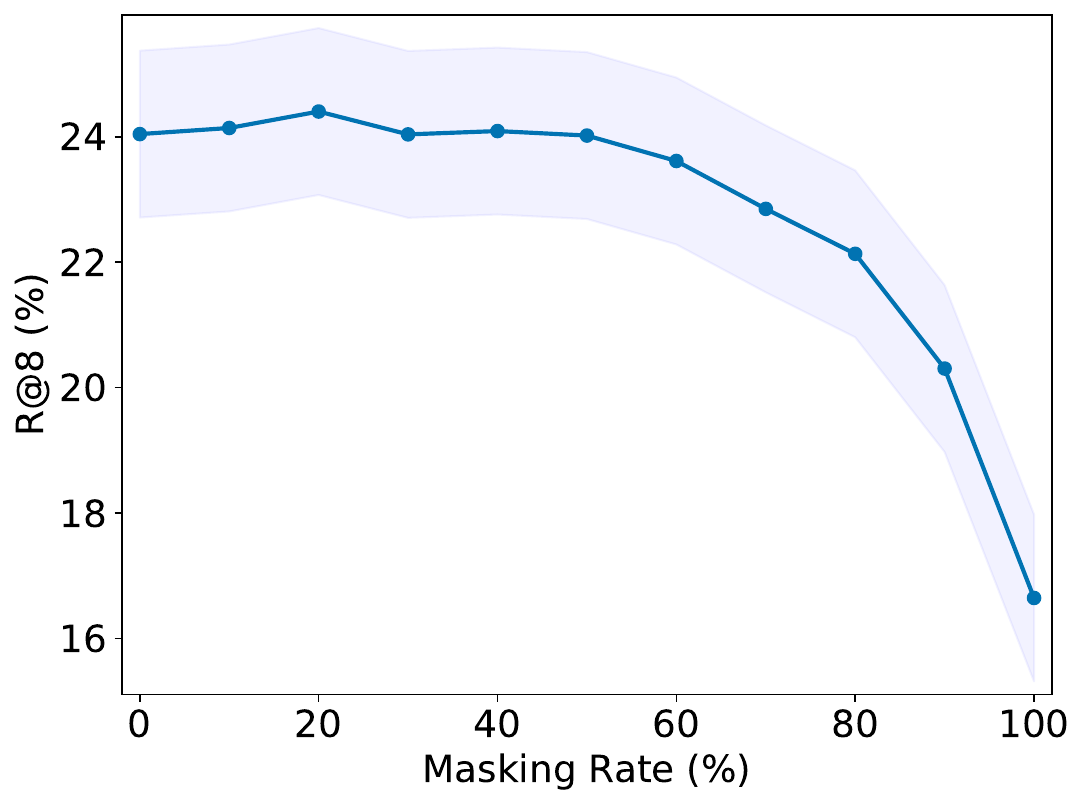}
    \caption{\textbf{Hyperparameter search identifies 20\% as the optimal masking rate for RoBERTa-large.}
    We search using a 10\% random subset of the training data and report R@8 on the validation set for seen English domains.}
    \label{fig:analysis1_maskrate_roberta}
\end{figure}

\begin{figure}[t]
    \centering
    \includegraphics[width=\columnwidth]{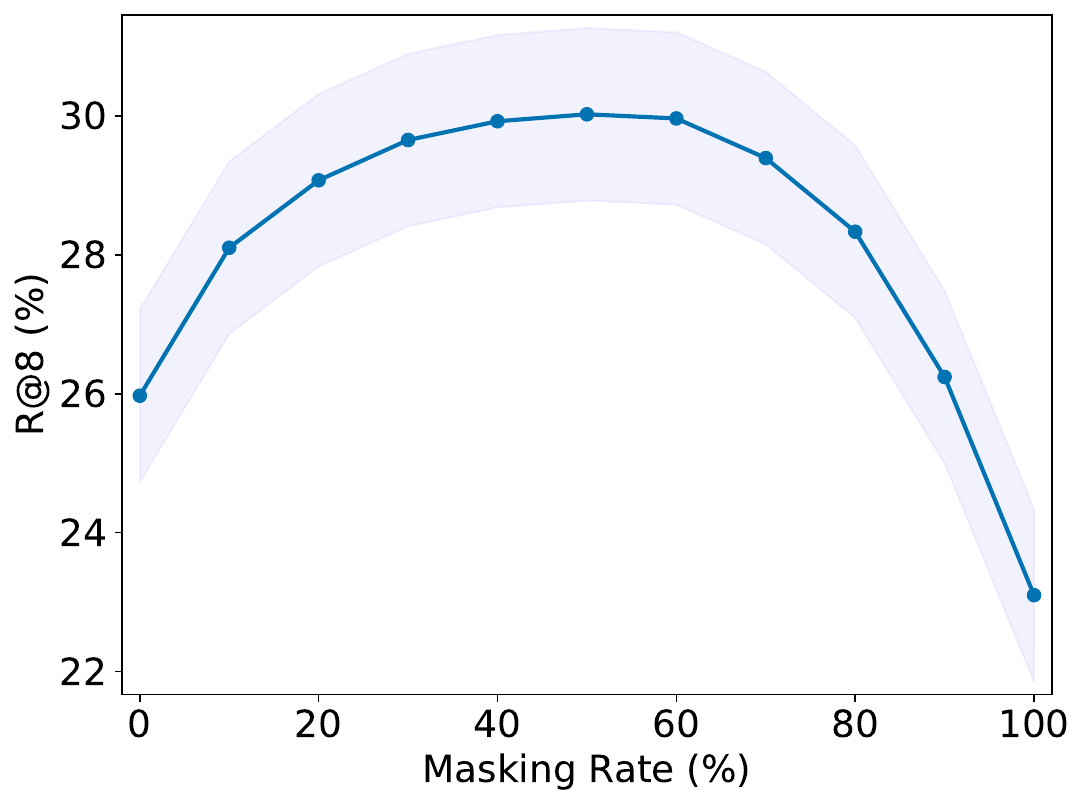}
    \caption{\textbf{Hyperparameter search identifies 50\% as the optimal masking rate for Llama3.2-1B.}
    We search using a 10\% random subset of the training data and report R@8 on the validation set for seen English domains.}
    \label{fig:analysis1_maskrate_llama}
\end{figure}

\section{Supplementary Material for Section~\ref{sec:analysis}}
We provide additional results of our analysis.

\subsection{Masking Rate Search}
The hyperparameter search results on XLM-RoBERTa-large, RoBERTa-large, and Llama3.2-1B are shown in Figure~\ref{fig:analysis1_maskrate_xlmroberta}, Figure~\ref{fig:analysis1_maskrate_roberta}, and Figure~\ref{fig:analysis1_maskrate_llama}, respectively.

\subsection{Ablation Study: PCM}
Table~\ref{tab:analysis2_pcm_full} includes Llama result in addition to Table~\ref{tab:analysis2_pcm}.
Table~\ref{tab:analysis2_pcm_mrr} presents the MRR results corresponding to Table~\ref{tab:analysis2_pcm_full}.

\subsection{Ablation Study: LAB}
Table~\ref{tab:analysis3_lab_mrr} presents the MRR results corresponding to Table~\ref{tab:analysis3_lab}.

\begin{table}[t]
\centering
\resizebox{\columnwidth}{!}{
    \begin{tabular}{llrr}
    \textbf{Model} & \textbf{Mask} & \textbf{Seen} & \textbf{Unseen} \\ 
    \hline
    RandomOrder & - & 0.0016  & 0.0023  \\
    \hline
    LUAR & \xmark & 10.35  & 11.79  \\
    CAV & \xmark & 1.50  & 2.75  \\
    StyleDistance & \xmark & 3.16  & 4.86  \\
    \hline
    \multirow{4}{*}{RoBERTa}
    & \xmark & 22.60  & 14.64  \\
    & Stopword & 10.34  & 7.88  \\
    & POS & 19.14  & 12.97  \\
    & PCM & \textbf{24.66} & \textbf{16.24} \\
    \hline
    \multirow{2}{*}{Llama}
    & \xmark & 28.11  & 17.69  \\
    & PCM & \textbf{29.00} & \textbf{19.36} \\
    \end{tabular}
}
\caption{\textbf{PCM achieves the highest R@8 among all evaluated methods on both seen and unseen English domains.}
It outperforms state-of-the-art AR methods by a substantial margin and surpasses masking baselines that rely on language-specific resources.}
\label{tab:analysis2_pcm_full}
\end{table}

\begin{table}[t]
\centering
\resizebox{\columnwidth}{!}{
    \begin{tabular}{llrr}
    \textbf{Model} & \textbf{Mask} & \textbf{Seen} & \textbf{Unseen} \\ 
    \hline
    RandomOrder & - & 0.0028  & 0.0038  \\
    \hline
    LUAR          & \xmark & 7.50 & 8.97 \\
    CAV           & \xmark & 1.16 & 2.17 \\
    StyleDistance & \xmark & 2.38 & 3.82 \\
    \hline
    \multirow{4}{*}{RoBERTa}
    & \xmark   & 16.43          & 10.93 \\
    & Stopword & 7.21           & 5.87 \\
    & PertLE   & 13.66          & 9.61 \\
    & PCM      & \textbf{18.08} & \textbf{12.20} \\
    \hline
    \multirow{2}{*}{Llama}
    & \xmark   & 21.07          & 13.37 \\
    & PCM      & \textbf{21.90} & \textbf{14.78} \\
    \end{tabular}
}
\caption{\textbf{PCM achieves the highest MRR among all evaluated methods on both seen and unseen English domains.}
It outperforms state-of-the-art AR methods by a substantial margin and surpasses masking baselines that rely on language-specific resources.}
\label{tab:analysis2_pcm_mrr}
\end{table}

\begin{table}[t]
\centering
\resizebox{\columnwidth}{!}{
    \begin{tabular}{lrrrr}
    \multirow{2}{*}{\textbf{LAB}} & \multicolumn{2}{c}{\textbf{English}} & \multicolumn{2}{c}{\textbf{Multilingual}} \\ 
    \cmidrule{2-5}
    & \textbf{Seen} & \textbf{Unseen} & \textbf{Seen} & \textbf{Unseen} \\ 
    \hline
    \xmark & \textbf{14.99} & 11.60          & 38.24          & 24.24 \\
    \cmark & 14.59          & \textbf{13.45} & \textbf{45.04} & \textbf{30.36} \\
    \end{tabular}
}
\caption{\textbf{LAB improves multilingual MRR and enhances both cross-lingual and cross-domain generalization in terms of MRR.}
These gains in unseen English domains and multilingual subsets come at only a minor cost in seen English domains.}
\label{tab:analysis3_lab_mrr}
\end{table}

\subsection{Authorship Verification}
Table~\ref{tab:analysis4_pan_av_full} shows the comprehensive per-split and overall authorship verification AUROC of all evaluated AR models.

\subsection{Machine-Generated Text Detection}
Table~\ref{tab:analysis5_multitude_ai_full} presents the comprehensive per-language and overall machine-generated text detection accuracy of all evaluated AR models.

\subsection{Case study}
We show examples of document pairs in Amazon domain with their authorship attribution results.

\paragraph{Ex1.}
\textit{Query}: great video it dose not look like other that is around the pirce it looks HD best buy for computer items in a long time use for video chat people that can teach me the things i need to know for hobbies like FCC radios skpye not to good go with ooVoo it free also\\
\textit{Candidate}: the photos are over 7mb each the space adds ups fast but the photos are great if you don't have much space on your hard drive on the camera you can change how many mp the photo take so the file size are little if you get this camera you should also get a 16 to 32gb sd card to hold all your photos\\
\textit{Rank}: 1, \textit{Attribution}: Success

\paragraph{Ex2.}
\textit{Query}: LOVE my Toms!  I was worried at first because they start off tight, and this was my first pair so I didn't know what to expect, but my sister told me to hang in there and wear for a couple days- and they definitely stretched out and fit my feet perfectly!  I've worn them on trips from the east to the west coast, and even over seas, and they are great!  So light for packing, comfy for walking, and easy to put on and take off.  They got muddy on a recent trip, but I let it dry and just brushed it off and they were fine. I have a couple pairs of their wedges, and plan on getting more of these as I love the comfort and their mission! I have the navy blue- and I wear them to work and casually, so they go great with a lot!\\
\textit{Candidate}: It's hard to find a long shower curtain, so this was a great find and is perfect!  High quality- I expect it will last for awhile, and it doesn't need a liner which is nice.  The colors are great, it doesn't stick to you (a nice soft fabric feel), and keeps everything in the shower.  It hangs great- you don't have to weigh down, and doesn't wrinkle.  We didn't use the rings it came with (they are cheaper plastic ones), but ones we already had are great. Highly recommend if you have a rod that is higher- we use in a stand-up shower.\\
\textit{Rank}: 1, \textit{Attribution}: Success

\paragraph{Ex3.}
\textit{Query}: Solid little interface for the money. I like it a lot. The only thing I think that could have been done better would be phantom power control for individual channels, and not just 1-2/3-4 grouping. I was hoping to run three condenser mics on my acoustic with the fourth input being for the plug, but to do this I will need to purchase a phantom-powered direct box. Not terrible, but not ideal either. Still a good solution for the home-studio budget musician.\\
\textit{Candidate}: I have a "Silent" PC build with a Core i7-5820k, a Noctua NH-D15, three additional case fans, DDR4 RAM, an ASUS X99 mobo, two SSDs, one HDD, and a GTX 970, and this provides enough power to my system that I have yet to actually hear the fan kick on.\\
\textit{Rank}: 9780, \textit{Attribution}: Fail

\paragraph{Ex4.}
\textit{Query}: THIS IS THE MOST COMFORTABLE HANG ON TYPE OF STAND ON EARTH!  Once you have hunted from a Millennium Tree Stand you will not be able to go back to ordinary. So purchase this stand with caution and have a yard sale for your now inadequate equipment. This is the first of the last stand you will ever buy.\\
\textit{Candidate}: This book has adorable illustrations and is written so that you can just read about one person at a time (you don't have to read the entire book in one setting).  A great way to introduce non-fiction to younger readers while also giving them a glimpse into strong women throughout history.\\
\textit{Rank}: 12018, \textit{Attribution}: Fail

\begin{table*}[t]
\centering
\large
\resizebox{\textwidth}{!}{ 
{\large
    \begin{tabular}{lrrrrrrr}
    \multirow{2}{*}{\textbf{Language}} & \multirow{2}{*}{\textbf{\#Data}} & \textbf{Monolingual} & \multirow{2}{*}{\textbf{mStyleDistance}} & \multicolumn{2}{c}{\textbf{XLM-RoBERTa}} & \multicolumn{2}{c}{\textbf{Llama3.2}}\\
    \cmidrule{5-8}
    & & \textbf{BERT} &  & \textbf{Monolingual}& \textbf{Multilingual} & \textbf{Monolingual} & \textbf{Multilingual}\\ 
    \hline
    French        & 8,944  & 60.14 & 5.58 & 55.48 & 55.52   & 60.58 & 58.47 \\
    Russian       & 5,584  & -     & 4.98 & 51.63 & 52.11 & 55.71 & 52.85 \\
    Italian       & 5,048  & 52.61 & 4.62 & 49.86 & 53.57   & 57.82 & 55.86 \\
    Polish        & 2,928  & 56.08 & 9.04 & 50.68 & 52.97    & 56.59 & 55.19 \\
    Arabic        & 2,412  & -     & 9.86 & 54.98 & 55.76 & 58.00 & 54.40 \\
    Farsi         & 2,192  & 51.09 & 8.17 & 51.23 & 55.38  & 58.71 & 55.47 \\
    Dutch         & 2,128  & 54.84 & 6.30 & 51.36 & 58.08    & 58.46 & 58.46 \\
    Japanese      & 1,752  & -     & 13.24 &52.45 & 51.77 & 59.36 & 55.25 \\
    Turkish       & 1,492  & -     & 7.64 & 53.89 & 57.44 & 58.45 & 57.53 \\
    Hebrew        & 1,372  & 56.12 & 7.43 & 51.82 & 59.11 & 48.40 & 40.64 \\
    Indonesian    & 1,268  & 26.18 & 9.94 & 56.31 & 59.62  & 61.83 & 60.05 \\
    Hungarian     & 1,200  & 56.50 & 7.90 & 57.58 & 61.08  & 59.58 & 60.33 \\
    Vietnamese    & 892   & -      & 9.83 &59.42 & 64.69 & 65.13 & 64.51 \\
    Greek         & 656   & -      & 9.57 &59.30 & 64.33 & 64.48 & 63.41 \\
    Serbian       & 644   & -      & 11.63 &63.20 & 65.53 & 64.91 & 62.97 \\
    Chinese       & 616   & -      & 15.70 & 49.19 & 52.44 & 54.55 & 54.38 \\
    Hindi         & 452   & -      & 16.34 &57.74 & 59.51 & 64.60 & 58.48 \\
    Thai          & 404   & -      & 18.72 &65.84 & 72.52 & 68.07 & 70.75 \\
    Macedonian    & 216   & -      & 13.76 &56.94 & 62.96 & 62.50 & 61.11 \\
    Kazakh        & 200   & -      & 11.39 &43.00 & 58.50 & 53.00 & 57.50 \\
    Georgian      & 176   & -      & 15.91 &56.82 & 72.73 & 57.95 & 48.30 \\
    Malayalam     & 172   & -      & 11.05 &54.70 & 63.95 & 42.44 & 36.31 \\
    \end{tabular} 
    }
}
\caption{Comparison between multilingual and monolingual training in terms of R@8.}
\label{tab:exp1_full_r@8}
\end{table*}

\begin{table*}[t]
\centering
\resizebox{\textwidth}{!}{ 
    \begin{tabular}{lrrrrrrr}
    \multirow{2}{*}{\textbf{Language}} & \multirow{2}{*}{\textbf{\#Data}}  & \textbf{Monolingual} & \multirow{2}{*}{\textbf{mStyleDistance}} & \multicolumn{2}{c}{\textbf{XLM-RoBERTa}} & \multicolumn{2}{c}{\textbf{Llama3.2}}\\
    \cmidrule{5-8}
    & & \textbf{BERT} &  & \textbf{Monolingual} & \textbf{Multilingual} & \textbf{Monolingual} & \textbf{Multilingual}\\ 
    \hline
    French        & 8,944  & 51.04  & 4.58 & 46.05 & 45.95  & 52.56 & 48.53 \\
    Russian       & 5,584  & -      & 3.76 & 42.37 & 41.47  & 46.28 & 42.26 \\
    Italian       & 5,048  & 41.60  & 3.58 & 39.86 & 42.93  & 48.03 & 44.83 \\
    Polish        & 2,928  & 46.64  & 7.75 & 42.07 & 43.68  & 47.59 & 45.06 \\
    Arabic        & 2,412  & -      & 8.25 &44.15 & 44.20  & 46.70 & 43.18 \\
    Farsi         & 2,192  & 40.43  & 6.96 & 41.89 & 44.67  & 48.40 & 44.27 \\
    Dutch         & 2,128  & 45.11  & 6.30 & 41.63 & 46.77  & 49.96 & 48.27 \\
    Japanese      & 1,752  & -      & 10.14 & 40.51 & 40.05  & 47.52 & 42.70 \\
    Turkish       & 1,492  & -      & 6.24 & 43.30 & 44.90  & 47.51 & 44.72 \\
    Hebrew        & 1,372  & 45.93  & 5.32 &41.62 & 47.52  & 36.35 & 28.29 \\
    Indonesian    & 1,268  & 16.46  & 6.94 & 46.72 & 48.61  & 50.66 & 47.50 \\
    Hungarian     & 1,200  & 44.12  & 6.29 & 45.31 & 48.36  & 48.69 & 48.00 \\
    Vietnamese    & 892   & -       & 7.64 &48.70 & 51.70  & 54.62 & 52.21 \\
    Greek         & 656   & -       & 6.98 &46.79 & 50.76  & 52.57 & 50.30 \\
    Serbian       & 644   & -       & 8.53 &49.49 & 54.84  & 52.70 & 51.53 \\
    Chinese       & 616   & -       & 12.35 &39.34 & 42.35  & 45.06 & 44.84 \\
    Hindi         & 452   & -       & 11.39 &45.48 & 46.77  & 51.71 & 44.55 \\
    Thai          & 404   & -       & 14.48 &50.95 & 56.76  & 54.20 & 55.03 \\
    Macedonian    & 216   & -       & 10.19 &42.27 & 49.61  & 48.93 & 46.32 \\
    Kazakh        & 200   & -       & 9.69 &30.78 & 45.52  & 37.68 & 40.60 \\
    Georgian      & 176   & -       & 10.54 &41.97 & 55.62  & 35.13 & 33.26 \\
    Malayalam     & 172   & -       & 7.33 &37.94 & 45.01  & 25.27 & 22.54 \\
    \end{tabular} 
}
\caption{Comparison between multilingual and monolingual training in terms of MRR.}
\label{tab:exp1_full_mrr}
\end{table*}

\clearpage
\begin{table*}[t]
\centering
\resizebox{\textwidth}{!}{
    \begin{tabular}{llrrrr}
    \multirow{2}{*}{\textbf{Model}} & \multirow{2}{*}{\textbf{Data}} & \multicolumn{2}{c}{\textbf{English}} & \multicolumn{2}{c}{\textbf{Multilingual}} \\ 
    \cmidrule{3-6}
    & & \textbf{Seen Domain} & \textbf{Unseen Domain} & \textbf{Seen Languages} & \textbf{Unseen Languages} \\ 
    \hline
    XLM-RoBERTa
    & English      & \textbf{15.30} & 11.02 & 34.39 & 22.44 \\
    & Multilingual & 14.96 & \textbf{11.55} & \textbf{39.20} & \textbf{24.95} \\
    \hline
    Llama
    & English      & \textbf{21.90} & 14.78 & 37.42 & 26.33 \\
    & Multilingual & 21.71 & \textbf{15.35} & \textbf{40.57} & \textbf{27.60} \\
    \end{tabular}
}
\caption{Comparison between models trained on multilingual dataset and English-only dataset in MRR metric.}
\label{tab:exp2_cross_lingual_domain_mrr}
\end{table*}

\begin{table*}[t]
\centering
\resizebox{\textwidth}{!}{
\begin{tabular}{l rr|rrr|rrr|r}
     \multirow{2}{*}{\textbf{Model}} & \multicolumn{2}{c}{\textbf{PAN 2013}} & \multicolumn{3}{c}{\textbf{PAN 2014}} & \multicolumn{3}{c|}{\textbf{PAN 2015}} & \multirow{2}{*}{\textbf{Overall}}\\
     \cmidrule{2-9}
     & \textbf{Greek} & \textbf{Spanish}& \textbf{Greek} & \textbf{Spanish} & \textbf{Dutch} & \textbf{Greek} & \textbf{Spanish} & \textbf{Dutch} & \\
     \hline
     LUAR & 50.22 & \textbf{89.10} & \textit{68.44} & 55.88 & 68.68 & \textit{73.40} & 58.92 & 35.95 & 56.48\\
     CAV & \textit{57.78} & \textit{87.82} & 54.68 & 56.28 & 62.92 & 58.88 & 60.68 & 33.24 & 53.28\\
     StyleDistance & 42.22 & 86.54 & 60.84 & \textbf{65.52} & 67.48 & 52.60 & 53.44 & \textbf{55.33} & 57.96\\
     mStyleDistance & \textbf{59.11} & 57.69 & 48.24 & 48.12 & 56.13 & 55.12 & 43.68 & 28.55 & 46.47\\
     Ours (XLM-R) & 52.89 & 69.23 & 67.68 & \textit{65.24} & \textbf{81.29} & \textbf{78.08} & \textbf{69.32} & 40.17 & \textbf{62.86}\\
     Ours (Llama) & 48.89 & 57.05 & \textbf{68.56} & 57.40 & \textit{78.08} & 70.24 & \textit{65.64} & \textit{40.38} & \textit{61.07}\\
\end{tabular}
}
\caption{The results of the PAN 2013-2015 AV shared task for Greek, Spanish, and Dutch are shown above.
Performance is reported separately for each PAN dataset, as well as the average performance across datasets for the same language.
Our method demonstrates the highest performance overall.
The best and the second-best performances are bolded and italicized, respectively.}
\label{tab:analysis4_pan_av_full}
\end{table*}

\begin{table*}[t]
\centering
\resizebox{\textwidth}{!}{
\begin{tabular}{l rrrrrrrrrrr|r}
    \multirow{2}{*}{\textbf{Model}} & \textbf{Arabic} & \textbf{Catalan} & \textbf{Chinese} & \textbf{Czech} & \textbf{Dutch} & \textbf{English} & \textbf{German} & \textbf{Portuguese} & \textbf{Russian} & \textbf{Spanish} & \textbf{Ukrainian} & \multirow{2}{*}{\textbf{Overall}}\\
     & \textbf{(ar)} & \textbf{(ca)} & \textbf{(zh)} & \textbf{(cs)} & \textbf{(nl)} & \textbf{(en)} & \textbf{(de)} & \textbf{(pt)} & \textbf{(ru)} & \textbf{(es)} & \textbf{(uk)} & \\
    \hline
    LUAR & \textit{88.81} & 84.17 & 77.90 & 87.10 & 83.93 & 91.37 & 88.45 & 76.77 & 89.93 & 90.62 & 80.32 & 85.36 \\
    CAV & 85.33 & 85.14 & 64.44 & 88.77 & 88.72 & 90.57 & 89.24 & 87.36 & 88.24 & 89.54 & 87.37 & 85.85\\
    StyleDistance & 87.65 & 88.41 & 69.44 & 44.85 & 81.56 & \textbf{95.18} & 64.25 & 85.97 & 89.70 & \textbf{93.31} & 88.19 & 80.66\\
    mStyleDistance & 86.68 & \textbf{88.93} & 86.73 & \textbf{90.48} & 88.61 & 91.89 & \textit{90.28} & 89.08 & 90.38 & \textit{91.63} & \textbf{89.17} & 89.42\\
    Ours (XLM-R) & 87.02 & 88.81 & \textbf{88.86} & \textit{89.92} & \textit{88.76} & \textit{93.34} & \textbf{91.55} & \textit{89.23} & \textbf{91.35} & 91.52 & \textit{88.87} & \textbf{89.91}\\
    Ours (Llama) & \textbf{88.85} & \textit{88.85} & \textit{88.74} & 88.84 & \textbf{88.91} & 92.53 & 89.94 & \textbf{89.30} & \textit{90.53} & 90.73 & \textit{88.87} & \textit{89.63}\\
\end{tabular}
}
\caption{The table displays the performance of the models on the MULTITuDE dataset for each language.
Our proposed model achieves the best overall performance.
The best and the second-best performances are bolded and italicized, respectively.}
\label{tab:analysis5_multitude_ai_full}
\end{table*}

\clearpage

\sisetup{
  group-separator = {,},
  group-minimum-digits = 4,
  table-format = 7.0,
  detect-weight = true
}

\begin{table*}[t]
\centering
\resizebox{0.85\textwidth}{!}{
    \begin{tabular}{lll}
    \multicolumn{2}{l}{\textbf{Domain}} & \textbf{Description} \\
    \hline
    \multicolumn{3}{l}{\textbf{Seen Domain}} \\
    & Reddit               & Entries in online forum Reddit \\
    & Gmane                & Emails from public mailing lists \\
    & StackExchange        & Entries in Q\&A community StackExchange \\
    & Wikipedia: article   & Articles in Wikipedia \\
    & AO3                  & Fan works from Archive of Our Own \\
    & RealNews             & News articles from Common Crawl \\
    & Amazon               & Reviews on Amazon \\
    & NYTimes              & Articles from The New York Times \\
    & BookCorpus           & Self-published novel books \\
    & PubMed               & Scientific publications from PubMed database \\
    \hline
    \multicolumn{3}{l}{\textbf{Unseen Domain}} \\
    & Goodreads            & Book reviews from website Goodreads \\
    & Wikipedia: talk      & Talks in Wikipedia \\
    & Wikipedia: user talk & User talks in Wikipedia \\
    & Wikipedia: user      & User pages in Wikipedia \\
    & food.com             & Food recipes from food.com \\
    & BlogCorpus           & Blog posts from blogger.com \\
    & SFU-SOCC             & Online news comments \\
    \end{tabular}
}
\caption{Domain descriptions for our dataset.}
\label{tab:setup_desc_english}
\end{table*}

\begin{table*}[t]
\centering
\resizebox{\textwidth}{!}{
\begin{tabular}{ll ll}
\multicolumn{2}{l}{\textbf{Language}} & \textbf{Family} & \textbf{Script} \\
\hline
\multicolumn{4}{l}{\textbf{Seen Language}} \\
& French    & Romance, Indo-European       & Latin\\
& Russian   & Slavic, Indo-European        & Cyrillic \\
& Italian   & Romance, Indo-European       & Latin\\
& Polish    & Slavic, Indo-European        & Latin \\
& Arabic    & Semitic, Afro-Asiatic        & Arabic \\
& Farsi     & Indo-Iranian, Indo-European  & Arabic (Persian variant)\\
& Dutch     & Germanic, Indo-European      & Latin \\
& Japanese  & Japonic                      & Kanji, Hiragana, Katakana \\
& Turkish   & Turkic                       & Latin\\
& Hebrew    & Semitic, Afro-Asiatic        & Hebrew \\
& Indonesian & Austronesian                & Latin \\
& Hungarian & Uralic                       & Latin\\
& Vietnamese & Austro-Asiatic              & Latin\\
& Greek     & Greek, Indo-European         & Greek \\
& Serbian   & Slavic, Indo-European        & Cyrillic \\
& Chinese   & Sino-Tibetan                 & Han \\
& Hindi     & Indo-Iranian, Indo-European  & Devanagari \\
& Thai      & Kra-Dai, Tai-Kadai           & Thai\\
& Macedonian & Slavic, Indo-European       & Cyrillic\\
& Kazakh    & Turkic                       & Cyrillic (formerly Latin and Arabic) \\
& Georgian  & Kartvelian                   & Georgian \\
& Malayalam & Dravidian                    & Malayalam\\
& Icelandic & Germanic, Indo-European      & Latin\\
& Swahili   & Atlantic-Congo, Niger-Congo  & Latin, Braille \\
& Punjabi   & Indo-Iranian, Indo-European  & Gurmukhi (Eastern Punjab), Shahmukhi (Western Punjab) \\
& Burmese   & Sino-Tibetan                 & Burmese \\
& Javanese  & Austronesian                 & Latin \\
& Gujarati  & Indo-Iranian, Indo-European  & Gujarati \\
& Hausa     & Chadic, Afro-Asiatic         & Latin \\
& Bengali   & Indo-Iranian, Indo-European  & Bengali \\
& Yoruba    & Atlantic-Congo, Niger-Congo  & Latin \\
& Amharic   & Semitic, Afro-Asiatic        & Ethiopic \\
& Malagasy  & Austronesian                 & Latin \\
& Chechen   & Nakh-Daghestanian            & Cyrillic \\
& Cherokee  & Iroquoian                    & Cherokee \\
\end{tabular}
}
\caption{The language family and script of seen languages in the multilingual Wikipedia portion of our dataset.}
\label{tab:setup_langfamily_multi_seen}
\end{table*}

\begin{table*}[t]
\centering
\resizebox{0.85\textwidth}{!}{
\begin{tabular}{l l l l}
\multicolumn{2}{l}{\textbf{Language}} & \textbf{Family} & \textbf{Script}\\
\hline
\multicolumn{4}{l}{\textbf{Unseen Language}} \\
& German     & Germanic, Indo-European    & Latin \\
& Spanish    & Romance, Indo-European     & Latin \\
& Portuguese & Romance, Indo-European     & Latin \\
& Ukrainian  & Slavic, Indo-European      & Cyrillic\\
& Czech      & Slavic, Indo-European      & Latin \\
& Swedish    & Germanic, Indo-European    & Latin \\
& Bulgarian  & Slavic, Indo-European      & Cyrillic \\
& Armenian   & Armenian, Indo-European    & Armenian\\
& Finnish    & Uralic                     & Latin\\
& Uzbek      & Turkic                     & Latin (formerly Cyrillic) \\
& Marathi    & Indo-Iranian, Indo-European & Devanagari \\
& Belarusian & Slavic, Indo-European      & Cyrillic\\
& Urdu       & Indo-Iranian, Indo-European & Arabic (Nastaliq variant)\\
& Telugu     & Dravidian                  & Telugu\\
& Tagalog    & Austronesian               & Latin\\
& Afrikaans  & Germanic, Indo-European    & Latin\\
& Egyptian Arabic & Semitic, Afro-Asiatic & Arabic (Naskh variant) \\
& Tatar      & Turkic                     & Cyrillic, Latin (formerly Arabic)\\
& Sundanese  & Austronesian               & Latin \\
& Zulu       & Atlantic-Congo, Niger-Congo & Latin\\
& Simple English & Germanic, Indo-European & Latin \\
& Mazanderani & Indo-Iranian, Indo-European & Arabic (Naskh, Nastaliq variant)\\
& Wu         & Sino-Tibetan                & Han\\
& Fulah      & Atlantic-Congo, Niger-Congo & Latin, Arabic\\
\end{tabular}
}
\caption{The language family and script of unseen languages in the multilingual Wikipedia portion of our dataset.}
\label{tab:setup_langfamily_multi_unseen}
\end{table*}

\begin{table*}[t]
\centering
\resizebox{0.85\textwidth}{!}{
\begin{tabular}{l l S S S S}
\multicolumn{2}{l}{\textbf{Domain}} & \textbf{Train} & \textbf{Validation} & \textbf{Test} & \textbf{Total} \\
\hline
\multicolumn{6}{l}{\textbf{Seen Domain}} \\
& Reddit & 2089782 & 122928 & 245858 & 2458568 \\
& Gmane & 536915 & 31583 & 63167 & 631665 \\
& StackExchange & 497071 & 29239 & 58480 & 584790 \\
& Wikipedia: article & 416872 & 24522 & 49044 & 490438 \\
& AO3 & 245595 & 14447 & 28894 & 288936 \\
& RealNews & 162852 & 9579 & 19160 & 191591 \\
& Amazon & 104464 & 6145 & 12290 & 122899 \\
& NYTimes & 100764 & 5927 & 11856 & 118547 \\
& BookCorpus & 55219 & 3248 & 6497 & 64964 \\
& PubMed & 5606 & 330 & 660 & 6596 \\
\hline
\multicolumn{2}{l}{\textbf{Total (Seen)}} & 4215140 & 247948 & 495906 & 4958994 \\
\hline
\multicolumn{6}{l}{\textbf{Unseen Domain}} \\
& Goodreads & {} & 53226 & 106454 & 159680 \\
& Wikipedia: talk & {} & 48809 & 97619 & 146428 \\
& Wikipedia: user talk & {} & 36334 & 72670 & 109004 \\
& Wikipedia: user & {} & 20353 & 40707 & 61060 \\
& food.com & {} & 6514 & 13030 & 19544 \\
& BlogCorpus & {} & 5594 & 11188 & 16782 \\
& SFU-SOCC & {} & 3957 & 7915 & 11872 \\
\hline
\multicolumn{2}{l}{\textbf{Total (Unseen)}} & {} & 174787 & 349583 & 524370 \\
\end{tabular}
}
\caption{The statistics on the number of authors across domains and splits for the English portion of our dataset.}
\label{tab:setup_stats_english}
\end{table*}


\begin{table*}[t]
\centering
\resizebox{0.85\textwidth}{!}{
\begin{tabular}{l l S S S S}
\multicolumn{2}{l}{\textbf{Language}} & \textbf{Train} & \textbf{Validation} & \textbf{Test} & \textbf{Total} \\
\hline
\multicolumn{6}{l}{\textbf{Seen Language}} \\
& French & 76019 & 4472 & 8944 & 89435 \\
& Russian & 47458 & 2791 & 5584 & 55833 \\
& Italian & 42898 & 2523 & 5048 & 50469 \\
& Polish & 24893 & 1464 & 2930 & 29287 \\
& Arabic & 20518 & 1207 & 2414 & 24139 \\
& Farsi & 18626 & 1095 & 2192 & 21913 \\
& Dutch & 18085 & 1064 & 2128 & 21277 \\
& Japanese & 14882 & 875 & 1752 & 17509 \\
& Turkish & 12676 & 745 & 1492 & 14913 \\
& Hebrew & 11654 & 685 & 1372 & 13711 \\
& Indonesian & 10774 & 634 & 1268 & 12676 \\
& Hungarian & 10210 & 600 & 1202 & 12012 \\
& Vietnamese & 7602 & 447 & 895 & 8944 \\
& Greek & 5585 & 328 & 658 & 6571 \\
& Serbian & 5477 & 322 & 645 & 6444 \\
& Chinese & 5242 & 308 & 618 & 6168 \\
& Hindi & 3845 & 226 & 453 & 4524 \\
& Thai & 3439 & 202 & 406 & 4047 \\
& Macedonian & 1844 & 108 & 218 & 2170 \\
& Kazakh & 1716 & 101 & 202 & 2019 \\
& Georgian & 1486 & 87 & 176 & 1749 \\
& Malayalam & 1460 & 86 & 172 & 1718 \\
& Icelandic & 764 & 45 & 91 & 900 \\
& Swahili & 563 & 33 & 67 & 663 \\
& Punjabi & 532 & 31 & 64 & 627 \\
& Burmese & 403 & 24 & 48 & 475 \\
& Javanese & 385 & 22 & 46 & 453 \\
& Gujarati & 290 & 17 & 35 & 342 \\
& Hausa & 232 & 14 & 28 & 274 \\
& Bengali & 150 & 9 & 18 & 177 \\
& Yoruba & 129 & 7 & 16 & 152 \\
& Amharic & 73 & 4 & 10 & 87 \\
& Malagasy & 42 & 2 & 6 & 50 \\
& Chechen & 11 & 1 & 2 & 14 \\
& Cherokee & 3 & 1 & 2 & 6 \\
\hline
\multicolumn{2}{l}{\textbf{Total (Seen)}} & 349966 & 20580 & 41202 & 411748 \\
\end{tabular}
}
\caption{The statistics on the number of authors across languages and splits for seen languages in the multilingual Wikipedia portion of our dataset.}
\label{tab:setup_stats_multi_seen}
\end{table*}

\begin{table*}[t]
\centering
\resizebox{0.85\textwidth}{!}{
\begin{tabular}{l l S S S}
\multicolumn{2}{l}{\textbf{Language}} & \textbf{Validation} & \textbf{Test} & \textbf{Total} \\
\hline
\multicolumn{4}{l}{\textbf{Unseen Language}} \\
& German & 37853 & 75708 & 113561 \\
& Spanish & 28499 & 56998 & 85497 \\
& Portuguese & 13665 & 27331 & 40996 \\
& Ukrainian & 6263 & 12528 & 18791 \\
& Czech & 4704 & 9408 & 14112 \\
& Swedish & 4371 & 8742 & 13113 \\
& Bulgarian & 1841 & 3684 & 5525 \\
& Armenian & 1641 & 3282 & 4923 \\
& Finnish & 976 & 1953 & 2929 \\
& Uzbek & 815 & 1631 & 2446 \\
& Marathi & 364 & 730 & 1094 \\
& Belarusian & 364 & 730 & 1094 \\
& Urdu & 361 & 724 & 1085 \\
& Telugu & 273 & 548 & 821 \\
& Tagalog & 211 & 423 & 634 \\
& Afrikaans & 162 & 325 & 487 \\
& Egyptian Arabic & 116 & 233 & 349 \\
& Tatar & 108 & 218 & 326 \\
& Sundanese & 73 & 147 & 220 \\
& Zulu & 21 & 42 & 63 \\
& Simple English & 20 & 42 & 62 \\
& Mazanderani & 14 & 29 & 43 \\
& Wu & 9 & 19 & 28 \\
& Fulah & 2 & 4 & 6 \\
\hline
\multicolumn{2}{l}{\textbf{Total (Seen)}} & 102726 & 205479 & 308205 \\
\end{tabular}
}
\caption{The statistics on the number of authors across languages and splits for unseen languages in the multilingual Wikipedia portion of our dataset.}
\label{tab:setup_stats_multi_unseen}
\end{table*}

\clearpage

\begin{table*}[t]
\centering
\resizebox{\textwidth}{!}{ 
    \begin{tabular}{lrrrrrrrr}
    \multirow{2}{*}{\textbf{Split}} & \multirow{2}{*}{\textbf{\#Data}} & \multicolumn{2}{c|}{\textbf{RoBERTa}} & \multicolumn{2}{c|}{\textbf{XLM-RoBERTa}} & \multicolumn{3}{c}{\textbf{Llama3.2}}\\
    \cmidrule{3-9}
    & & \textbf{Eng(Ours)} & \multicolumn{1}{r|}{\textbf{Eng(No Mask)}} & \textbf{Eng(Ours)} & \multicolumn{1}{r|}{\textbf{Multi(Ours)}} & \textbf{Eng(Ours)} & \textbf{Eng(No Mask)} & \textbf{Multi(Ours)}\\ 
    \hline
    \multicolumn{9}{l}{\textbf{Seen Domain}}\\
    Reddit        & 245,858 & 8.21 & 7.12 & 5.98 & 5.35 & 9.69 & 8.85 & 9.28\\
    Gmane         & 63,167 & 51.54 & 47.92 & 43.53 & 40.77 & 62.18 & 62.12 & 59.50\\
    StackExchange & 58,480 & 15.03 & 13.19 & 13.28 & 12.04 & 22.32 & 20.33 & 21.42\\
    Wikipedia: article & 49,044 & 52.80 & 49.77 & 49.20 & 50.25 & 58.69 & 58.98 & 58.31\\
    AO3           & 28,894 & 53.45 & 49.45 & 48.30 & 44.73 & 59.63 & 58.19 & 56.52\\
    RealNews      & 19,160 & 56.28 & 52.38 & 48.63 & 46.22 & 62.45 & 61.87 & 60.16\\
    Amazon        & 12,290 & 17.94 & 16.09 & 14.43 & 13.56 & 20.48 & 18.42 & 19.54\\
    NYTimes       & 11,856 & 17.92 & 16.39 & 13.70 & 13.62 & 20.75 & 18.91 & 20.31\\
    BookCorpus    & 6,497 & 61.64 & 56.90 & 55.38 & 52.95 & 63.35 & 62.09 & 65.14\\
    PubMed        & 660 & 74.92 & 71.36 & 66.82 & 68.33 & 85.21 & 83.54 & 84.91\\
    \hline
    \multicolumn{9}{l}{\textbf{Unseen Domain}}\\
    Goodreads     & 106,454 & 4.85 & 4.06 & 6.21 & 6.10 & 5.16 & 4.68 & 7.03\\
    Wikipedia: talk & 97,619 & 22.84 & 20.27 & 19.01 & 24.03 & 28.60 & 26.33 & 32.00\\
    Wikipedia: user talk & 72,670 & 22.11 & 20.19 & 19.46 & 26.52 & 27.31 & 25.25 & 32.42\\
    Wikipedia: user & 40,707 & 33.08 & 31.73 & 32.01 & 33.84 & 36.82 & 36.42 & 37.99\\
    food.com      & 13,030 & 2.90 & 2.28 & 2.26 & 2.92 & 2.82 & 2.03 & 2.91\\
    BlogCorpus    & 11,188 & 36.83	& 33.59 & 28.17	& 29.10 & 34.33	& 31.45 & 36.66\\
    SFU-SOCC      & 7,915 & 13.02 & 11.78 & 13.06 & 15.30 & 15.22 & 12.54 & 17.64\\
    \end{tabular} 
}
\caption{Comprehensive per-domain R@8 results for multilingual and English-only models.}
\label{tab:full_english_r@8}
\end{table*}

\begin{table*}[t]
\centering
\large
\resizebox{\textwidth}{!}{ 
{\large
    \begin{tabular}{lrrrrrrrr}
    \multirow{2}{*}{\textbf{Split}} & \multirow{2}{*}{\textbf{\#Data}} & \multicolumn{2}{c|}{\textbf{RoBERTa}} & \multicolumn{2}{c|}{\textbf{XLM-RoBERTa}} & \multicolumn{3}{c}{\textbf{Llama3.2}}\\
    \cmidrule{3-9}
    & & \textbf{Eng(Ours)} & \multicolumn{1}{r|}{\textbf{Eng(No Mask)}} & \textbf{Eng(Ours)} & \multicolumn{1}{r|}{\textbf{Multi(Ours)}} & \textbf{Eng(Ours)} & \textbf{Eng(No Mask)} & \textbf{Multi(Ours)}\\ 
    \hline
    \multicolumn{9}{l}{\textbf{Seen Domain}}\\
    Reddit        & 245,858 & 5.34 & 4.60 & 3.88 & 3.51 & 6.35 & 5.70 & 6.12\\
    Gmane         & 63,167 & 38.88 & 35.93 & 32.29 & 30.04 & 48.91 & 49.02 & 46.55\\
    StackExchange & 58,480 & 9.65 & 8.49 & 8.52 & 7.92 & 15.19 & 13.51 & 14.56\\
    Wikipedia: article & 49,044 & 42.30 & 39.24 & 39.00 & 40.42 & 48.66 & 48.33 & 48.79\\
    AO3           & 28,894 & 40.81 & 37.25 & 36.41 & 33.47 & 46.97 & 44.56 & 44.31\\
    RealNews      & 19,160 & 40.96 & 37.51 & 35.00 & 33.64 & 46.72 & 45.70 & 45.13\\
    Amazon        & 12,290 & 11.41 & 10.15 & 9.33 & 8.73 & 13.09 & 11.85 & 12.65\\
    NYTimes       & 11,856 & 11.64 & 10.53 & 9.08 & 9.11 & 13.58 & 12.10 & 13.54\\
    BookCorpus    & 6,497 & 44.52 & 39.81 & 39.91 & 37.78 & 48.39 & 47.81 & 50.02\\
    PubMed        & 660 & 57.84 & 53.16 & 47.59 & 49.25 & 71.80 & 65.78 & 70.20\\
    \hline
    \multicolumn{9}{l}{\textbf{Unseen Domain}}\\
    Goodreads     & 106,454 & 3.03 & 2.56 & 3.94 & 3.88 & 3.29 & 3.00 & 4.50\\
    Wikipedia: talk & 97,619 & 16.88 & 14.68 & 13.87 & 17.96 & 21.74 & 19.58 & 24.89\\
    Wikipedia: user talk & 72,670 & 17.08 & 15.53 & 15.00 & 20.90 & 21.57 & 19.89 & 26.21\\
    Wikipedia: user & 40,707 & 27.35 & 26.10 & 26.61 & 28.15 & 30.60 & 30.02 & 31.55\\
    food.com      & 13,030 & 2.02 & 1.58 & 1.78 & 2.21 & 1.86 & 1.45 & 2.01\\
    BlogCorpus    & 11,188 & 26.05 & 23.54 & 19.27 & 20.06 & 24.84 & 21.81 & 26.63\\
    SFU-SOCC      & 7,915 & 8.67 & 7.62 & 8.79 & 10.27 & 10.00 & 8.49 & 11.96\\
    \end{tabular} 
    }
}
\caption{Comprehensive per-domain MRR results for multilingual and English-only models.}
\label{tab:full_english_mrr}
\end{table*}

\begin{table*}[t]
\centering
\resizebox{\textwidth}{!}{ 
    \begin{tabular}{lrrrrrrrr}
    \multirow{2}{*}{\textbf{Split}} & \multirow{2}{*}{\textbf{\#Data}} & \multicolumn{2}{c|}{\textbf{RoBERTa}} & \multicolumn{2}{c|}{\textbf{XLM-RoBERTa}} & \multicolumn{3}{c}{\textbf{Llama3.2}}\\
    \cmidrule{3-9}
    & & \textbf{Eng(Ours)} & \multicolumn{1}{r|}{\textbf{Eng(No Mask)}} & \textbf{Eng(Ours)} & \multicolumn{1}{r|}{\textbf{Multi(Ours)}} & \textbf{Eng(Ours)} & \textbf{Eng(No Mask)} & \textbf{Multi(Ours)}\\ 
    \hline
    \multicolumn{9}{l}{\textbf{Seen Language}}\\
    French     & 8,944 & 35.64 & 30.65 & 43.82 & 55.52 & 51.02 & 49.76 & 58.47 \\
    Russian    & 5,584 & 7.17 & 6.56 & 42.10 & 52.11 & 45.49 & 43.95 & 52.85 \\
    Italian    & 5,048 & 33.76 & 27.93 & 41.03 & 53.57 & 47.61 & 46.88 & 55.86 \\
    Polish     & 2,930 & 30.21 & 26.16 & 43.82 & 52.97 & 46.82 & 45.77 & 55.19 \\
    Arabic     & 2,414 & 11.46 & 8.96 & 43.95 & 55.76 & 41.07 & 41.89 & 54.40 \\
    Farsi      & 2,192 & 9.22 & 7.53 & 45.57 & 55.38 & 43.07 & 41.97 & 55.47 \\
    Dutch      & 2,128 & 37.95 & 32.50 & 46.43 & 58.08 & 51.88 & 49.44 & 58.46 \\
    Japanese   & 1,752 & 20.52 & 17.64 & 45.21 & 51.77 & 50.17 & 49.25 & 55.25 \\
    Turkish    & 1,492 & 27.78 & 23.66 & 48.53 & 57.44 & 51.68 & 49.13 & 57.53 \\
    Hebrew     & 1,372 & 10.17 & 8.05 & 48.98 & 59.11 & 18.20 & 21.03 & 40.64 \\
    Indonesian & 1,268 & 34.62 & 31.07 & 51.81 & 59.62 & 51.11 & 49.60 & 60.05 \\
    Hungarian  & 1,202 & 27.21 & 23.46 & 48.83 & 61.08 & 49.08 & 49.83 & 60.33 \\
    Vietnamese & 895 & 27.58 & 21.92 & 58.52 & 64.69 & 57.14 & 58.48 & 64.51 \\
    Greek      & 658 & 16.46 & 13.26 & 59.45 & 64.33 & 57.32 & 54.57 & 63.41 \\
    Serbian    & 645 & 20.11 & 17.62 & 56.52 & 65.53 & 56.56 & 55.63 & 62.97 \\
    Chinese    & 618 & 23.21 & 20.54 & 47.56 & 52.44 & 49.84 & 47.53 & 54.38 \\
    Hindi      & 453 & 14.27 & 12.83 & 51.55 & 59.51 & 49.11 & 47.32 & 58.48 \\
    Thai       & 406 & 20.30 & 14.98 & 62.87 & 72.52 & 63.25 & 64.50 & 70.75 \\
    Macedonian & 218 & 19.91 & 18.75 & 54.63 & 62.96 & 54.17 & 53.37 & 61.11 \\
    Kazakh     & 202 & 24.25 & 20.00 & 51.00 & 58.50 & 43.50 & 44.79 & 57.50 \\
    Georgian   & 176 & 20.74 & 15.06 & 67.05 & 72.73 & 26.14 & 28.41 & 48.30 \\
    Malayalam  & 172 & 11.92 & 13.08 & 51.16 & 63.95 & 19.64 & 23.13 & 36.31 \\
    Icelandic  & 91 & 36.36 & 33.52 & 63.64 & 80.68 & 59.09 & 57.50 & 76.14 \\
    Swahili    & 67 & 44.53 & 45.31 & 59.38 & 64.06 & 46.88 & 56.25 & 60.94 \\
    Punjabi    & 64 & 32.03 & 25.78 & 60.94 & 67.19 & 34.38 & 28.13 & 43.75 \\
    Burmese    & 48 & 34.38 & 32.29 & 75.00 & 83.33 & 39.58 & 41.67 & 47.92 \\
    Javanese   & 46 & 53.41 & 46.59 & 65.91 & 56.82 & 57.50 & 45.83 & 62.50 \\
    Gujarati   & 35 & 37.50 & 25.00 & 68.75 & 90.63 & 37.50 & 37.50 & 46.88 \\
    Hausa      & 28 & 71.43 & 62.50 & 67.86 & 78.57 & 87.50 & 87.50 & 95.83 \\
    Bengali    & 18 & 53.13 & 62.50 & 81.25 & 68.75 & 87.50 & 75.00 & 75.00 \\
    Yoruba     & 16 & 87.50 & 81.25 & 87.50 & 93.75 & 87.50 & 87.50 & 87.50 \\
    Amharic    & 10 & 100.00 & 100.00 & 100.00 & 100.00	 & 100.00 & 100.00 & 100.00\\
    Malagasy   & 6 & 100.00 & 100.00 & 100.00 & 100.00 & 100.00 & 100.00 & 100.00\\
    \hline
    \multicolumn{9}{l}{\textbf{Unseen Language}}\\
    German         & 75,708 & 17.73 & 13.44 & 25.01 & 34.25 & 31.02 & 29.35 & 35.50 \\
    Spanish        & 56,998 & 24.31 & 19.41 & 31.15 & 39.53 & 37.64 & 35.25 & 42.29 \\
    Portuguese     & 27,331 & 25.57 & 20.65 & 35.85 & 44.49 & 39.68 & 38.03 & 44.84 \\
    Ukrainian      & 12,528 & 5.35 & 4.65 & 31.63 & 40.77 & 30.18 & 31.19 & 38.94 \\
    Czech          & 9,408 & 20.90 & 16.95 & 36.61 & 46.95 & 41.40 & 38.63 & 46.39 \\
    Swedish        & 8,742 & 28.13 & 23.56 & 40.07 & 50.77 & 45.74 & 43.27 & 50.01 \\
    Bulgarian      & 3,684 & 8.06 & 7.13 & 41.99 & 52.01 & 41.11 & 39.70 & 46.68 \\
    Armenian       & 3,282 & 5.87 & 3.86 & 25.03 & 29.30 & 9.54 & 9.39 & 13.29 \\
    Finnish        & 1,953 & 13.14 & 11.01 & 21.06 & 23.51 & 21.47 & 20.75 & 24.39 \\
    Uzbek          & 1,631 & 18.06 & 15.60 & 33.60 & 36.18 & 33.95 & 30.94 & 37.13 \\
    Marathi        & 730 & 11.47 & 10.85 & 39.84 & 48.08 & 40.52 & 37.50 & 41.21 \\
    Belarusian     & 730 & 11.54 & 9.82 & 45.05 & 51.51 & 40.93 & 40.42 & 50.69 \\
    Urdu           & 724 & 10.64 & 10.36 & 44.34 & 52.62 & 35.69 & 33.06 & 46.94 \\
    Telugu         & 548 & 9.03 & 5.93 & 41.06 & 49.27 & 14.34 & 15.81 & 20.40 \\
    Tagalog        & 423 & 44.17 & 42.14 & 54.76 & 61.67 & 54.72 & 53.61 & 59.43 \\
    Afrikaans      & 325 & 43.21 & 38.43 & 51.54 & 58.95 & 48.13 & 48.13 & 51.88 \\
    Egyptian Arabic & 233 & 23.92 & 18.75 & 62.07 & 74.14 & 60.78 & 62.05 & 72.41 \\
    Tatar          & 218 & 20.83 & 18.98 & 32.41 & 37.50 & 43.52 & 36.54 & 50.93 \\
    Sundanese      & 147 & 28.47 & 27.43 & 45.14 & 52.78 & 40.97 & 39.58 & 43.75 \\
    Zulu           & 42 & 56.25 & 53.75 & 77.50 & 72.50 & 65.00 & 59.38 & 70.00 \\
    Simple English & 42 & 93.75 & 95.00 & 92.50 & 95.00 & 92.50 & 100.00 & 92.50 \\
    Mazanderani    & 29 & 58.93 & 53.57 & 75.00 & 82.14 & 83.33 & 50.00 & 95.83 \\
    Wu             & 19 & 62.50 & 65.63 & 81.25 & 75.00 & 68.75 & 68.75 & 75.00 \\
    Fulah          & 4 & 100.00 & 100.00 & 100.00 & 100.00 & 100.00 & 100.00 & 100.00\\
    \end{tabular} 
}
\caption{Comprehensive per-language R@8 results for multilingual and English-only models.}
\label{tab:full_multi_r@8}
\end{table*}


\begin{table*}[t]
\centering
\large
\resizebox{\textwidth}{!}{ 
{\large
    \begin{tabular}{lrrrrrrrr}
    \multirow{2}{*}{\textbf{Split}} & \multirow{2}{*}{\textbf{\#Data}} & \multicolumn{2}{c|}{\textbf{RoBERTa}} & \multicolumn{2}{c|}{\textbf{XLM-RoBERTa}} & \multicolumn{3}{c}{\textbf{Llama3.2}}\\
    \cmidrule{3-9}
    & & \textbf{Eng(Ours)} & \multicolumn{1}{r|}{\textbf{Eng(No Mask)}} & \textbf{Eng(Ours)} & \multicolumn{1}{r|}{\textbf{Multi(Ours)}} & \textbf{Eng(Ours)} & \textbf{Eng(No Mask)} & \textbf{Multi(Ours)}\\ 
    \hline
    \multicolumn{9}{l}{\textbf{Seen Language}}\\
    French     & 8,944 & 26.73 & 22.24 & 34.45 & 45.95 & 41.09 & 39.68 & 48.53\\
    Russian    & 5,584 & 5.36 & 4.75 & 32.21 & 41.47 & 35.45 & 33.21 & 42.26\\
    Italian    & 5,048 & 24.85 & 19.95 & 32.04 & 42.93 & 37.61 & 35.99 & 44.83\\
    Polish     & 2,930 & 22.58 & 19.66 & 34.37 & 43.68 & 38.17 & 35.88 & 45.06\\
    Arabic     & 2,414 & 9.44 & 7.02 & 34.29 & 44.20 & 31.82 & 31.94 & 43.18\\
    Farsi      & 2,192 & 7.46 & 6.08 & 35.20 & 44.67 & 33.26 & 31.56 & 44.27\\
    Dutch      & 2,128 & 28.81 & 23.87 & 36.62 & 46.77 & 41.43 & 39.84 & 48.27\\
    Japanese   & 1,752 & 14.99 & 12.60 & 34.27 & 40.05 & 38.93 & 37.10 & 42.70\\
    Turkish    & 1,492 & 19.70 & 16.77 & 36.59 & 44.90 & 39.20 & 36.71 & 44.72\\
    Hebrew     & 1,372 & 7.58 & 5.84 & 38.42 & 47.52 & 13.12 & 15.01 & 28.29\\
    Indonesian & 1,268 & 25.11 & 22.12 & 40.33 & 48.61 & 38.39 & 37.71 & 47.50\\
    Hungarian  & 1,202 & 19.49 & 16.07 & 38.28 & 48.36 & 38.18 & 37.25 & 48.00\\
    Vietnamese & 895 & 19.46 & 16.20 & 46.08 & 51.70 & 46.30 & 44.83 & 52.21\\
    Greek      & 658 & 11.65 & 9.21 & 44.91 & 50.76 & 43.65 & 43.41 & 50.30\\
    Serbian    & 645 & 14.81 & 13.03 & 44.34 & 54.84 & 44.48 & 42.49 & 51.53\\
    Chinese    & 618 & 17.52 & 14.35 & 38.99 & 42.35 & 39.89 & 38.71 & 44.84\\
    Hindi      & 453 & 10.57 & 9.61 & 39.97 & 46.77 & 36.92 & 37.18 & 44.55\\
    Thai       & 406 & 15.00 & 11.49 & 52.03 & 56.76 & 47.84 & 50.31 & 55.03\\
    Macedonian & 218 & 13.12 & 11.99 & 42.45 & 49.61 & 41.15 & 38.40 & 46.32\\
    Kazakh     & 202 & 16.52 & 12.76 & 38.91 & 45.52 & 31.34 & 30.22 & 40.60\\
    Georgian   & 176 & 15.84 & 10.15 & 48.78 & 55.62 & 20.33 & 19.39 & 33.26\\
    Malayalam  & 172 & 8.31 & 7.30 & 34.53 & 45.01 & 13.97 & 14.27 & 22.54\\
    Icelandic  & 91 & 23.46 & 21.39 & 48.39 & 57.15 & 33.60 & 39.86 & 52.05\\
    Swahili    & 67 & 33.48 & 30.87 & 43.75 & 42.18 & 39.12 & 41.54 & 38.28\\
    Punjabi    & 64 & 18.84 & 14.77 & 47.02 & 51.22 & 18.13 & 19.27 & 27.96\\
    Burmese    & 48 & 21.98 & 21.23 & 58.61 & 64.08 & 25.88 & 21.79 & 32.17\\
    Javanese   & 46 & 35.35 & 29.49 & 48.41 & 46.65 & 33.33 & 39.06 & 46.61\\
    Gujarati   & 35 & 20.20 & 13.27 & 52.31 & 54.96 & 23.61 & 20.13 & 26.57\\
    Hausa      & 28 & 45.47 & 45.89 & 55.35 & 54.21 & 53.17 & 58.41 & 52.48\\
    Bengali    & 18 & 36.46 & 31.67 & 46.28 & 42.61 & 51.27 & 45.00 & 58.51\\
    Yoruba     & 16 & 45.68 & 44.03 & 52.33 & 50.12 & 56.15 & 45.72 & 61.89\\
    Amharic    & 10 & 24.30 & 31.32 & 60.73 & 84.38 & 37.72 & - & 53.35\\
    Malagasy   & 6 & 100.00 & 93.75 & 87.50 & 100.00 & 100.00 & 100.00 & 100.00\\
    \hline
    \multicolumn{9}{l}{\textbf{Unseen Language}}\\
    German         & 75,708 & 12.82 & 9.61 & 18.75 & 26.63 & 23.47 & 22.04 & 27.36 \\
    Spanish        & 56,998 & 17.70 & 14.03 & 23.42 & 30.68 & 28.53 & 26.54 & 32.87\\
    Portuguese     & 27,331 & 18.47 & 14.72 & 26.77 & 34.47 & 0.13 & 28.50 & 34.71\\
    Ukrainian      & 12,528 & 4.17 & 3.51 & 22.89 & 31.06 & 21.99 & 22.46 & 29.21\\
    Czech          & 9,408 & 14.81 & 11.81 & 26.93 & 35.77 & 31.19 & 28.64 & 6.25\\
    Swedish        & 8,742 & 20.63 & 16.99 & 30.53 & 40.73 & 35.43 & 33.40 & 39.47\\
    Bulgarian      & 3,684 & 5.70 & 5.14 & 30.85 & 39.94 & 30.22 & 28.81 & 35.47\\
    Armenian       & 3,282 & 4.44 & 3.23 & 18.37 & 21.60 & 6.95 & 7.11 & 9.55\\
    Finnish        & 1,953 & 8.93 & 7.24 & 14.21 & 15.77 & 14.79 & 14.04 & 16.72\\
    Uzbek          & 1,631 & 12.23 & 10.61 & 23.61 & 26.17 & 23.34 & 21.64 &	26.46\\
    Marathi        & 730 & 8.36 &	7.99 & 29.12 & 35.87 & 27.85 & 25.95 & 29.71\\
    Belarusian     & 730 & 7.73 & 6.17 & 33.20 & 39.53 & 29.70 & 28.55 & 37.28\\
    Urdu           & 724 & 7.70 & 6.77 & 32.93 & 40.87 & 25.46 & 23.58 & 34.83\\
    Telugu         & 548 & 6.84 & 4.80 & 27.93 & 36.41 & 9.61 & 10.08 & 13.98\\
    Tagalog        & 423 & 31.74 & 30.00 & 41.29 & 48.82 & 40.38 & 0.31 & 5.91\\
    Afrikaans      & 325 & 31.72 & 27.84 & 40.10 & 45.80 & 39.35 & 39.82 & 40.97\\
    Egyptian Arabic & 233 & 18.64 & 13.75 & 51.16 & 58.24 & 49.85 & 49.36 & 58.19\\
    Tatar          & 218 & 13.43 & 10.97 & 21.76 & 25.71 & 28.08 & 25.79 & 33.79\\
    Sundanese      & 147 & 21.84 & 20.71 & 31.79 & 37.17 & 31.70 & 28.42 & 34.07\\
    Zulu           & 42 & 44.73 & 41.90 & 55.23 & 53.81 & 43.28 & 41.05 & 45.32\\
    Simple English & 42 & 82.22 & 80.48 & 80.03 & 80.29 & 83.01 & 91.02 & 83.48\\
    Mazanderani    & 29 & 35.27 & 28.71 & 59.46 & 66.35 & 63.80 & 45.01 & 71.18\\
    Wu             & 19 & 27.47 & 30.52 & 50.32 & 56.88 & 47.14 & 52.21 & 51.26\\
    Fulah          & 4 & 91.67 & 93.75 & 100.00 & 100.00 & 100.00 & 100.00 & 100.00\\
    \end{tabular} 
    }
}
\caption{Comprehensive per-language MRR results for multilingual and English-only models.}
\label{tab:full_multi_mrr}
\end{table*}

\end{document}